\documentclass[twoside,11pt]{article}
\usepackage{jair, theapa, rawfonts}

\usepackage{amsmath}
\DeclareMathOperator*{\argmax}{argmax}
\DeclareMathOperator*{\argmin}{argmin}
\newcommand{\hist}{\mbox{$\mathbf{y}_{1:j-1}$}}
\usepackage{amsfonts}
\usepackage{multirow}
\usepackage{graphicx}
\usepackage[format=plain]{caption}

\usepackage{bm}
\usepackage{array}
\newcolumntype{P}[1]{>{\centering\arraybackslash}p{#1}}
\usepackage{tablefootnote}

\ShortHeadings{Domain Adaptation for Neural Machine Translation: A Survey}{Danielle Saunders}

\begin{document}
\title{Domain Adaptation and Multi-Domain Adaptation for Neural Machine Translation: A Survey}
\author{\name Danielle Saunders \email ds636@cantab.ac.uk \\  \addr University of Cambridge, Engineering Department\\ Cambridge, United Kingdom}

\maketitle

\begin{abstract}
The development of deep learning techniques has allowed Neural Machine Translation (NMT) models to become extremely powerful, given sufficient training data and training time. However, systems struggle when translating text from a new domain with a distinct style or vocabulary.  Fine-tuning on in-domain data allows good domain adaptation, but requires sufficient relevant bilingual data. Even if this  is available, simple fine-tuning can cause overfitting to new data and `catastrophic forgetting' of previously learned behaviour.  

We concentrate on robust approaches to domain adaptation for NMT, particularly  where a system may need to translate across multiple domains. We divide techniques into those revolving around data selection or generation, model architecture,  parameter adaptation procedure, and inference procedure. We finally highlight the benefits of domain adaptation and multi-domain adaptation techniques to other lines of NMT research.

\end{abstract}
\section{Introduction}
\label{Introduction}

Neural Machine Translation (NMT) has seen impressive advances for some translation tasks in recent years. News and biomedical translation shared tasks from the Conference on Machine Translation (WMT) in 2019 already identified several systems as performing on par with a human translator for some high-resource language pairs according to human judgements \shortcite{barrault-etal-2019-findings,bawden-etal-2019-findings}.  Indeed, these tasks involve not only high-resource language pairs but also relatively high-resource domains, with millions of relevant sentence pairs available for training. However, NMT models perform less well on out-of-domain data. A model trained on exclusively news data is unlikely to achieve good performance on the biomedical domain, let alone human parity. 

Models trained on data from \emph{all} domains of interest can perform well across these domains. However, there is always the possibility of additional domains becoming interesting at a later stage. For example,  generic news translation systems of 2019 do not perform optimally on news from 2020 which heavily reports on the coronavirus pandemic \shortcite{anastasopoulos-etal-2020-tico}. While it is  possible to train a new model across all datasets from scratch for every new domain of interest, it is not generally practical, with typical training sets running to millions of sentence pairs and requiring multiple days for training \shortcite{junczys-dowmunt-2019-microsoft}. General overviews of NMT tend to focus on the many design decisions involved in this from-scratch NMT system development \shortcite{stahlberg2020neural,koehn2017nmt,neubig2017nmt}. However, once such a system is trained it can be repurposed to translate new types of language far faster than training a new model, and often performs better on the language of interest as a result  \shortcite{luong-etal-2015-addressing,freitag2016fast}.  This repurposing is generally known as domain adaptation.  

We interpret domain adaptation as any scheme intended to improve translations from an existing system for a certain topic or genre of language. In the definition of a domain we primarily follow \shortciteA{koehn-knowles-2017-six}, who state that a domain `may differ from other domains in topic, genre, style, level of formality, etc.'. However, we add some provisos when defining a domain as `a corpus from a specific source', often termed domain provenance. Provenance can be useful for describing adaptation, for example if applying in-domain translation to a fixed set of sentences. However, we do not treat it as an exclusive domain marker for the following reasons:

\begin{itemize}
    \item The provenance of unseen test data may not be known.
    \item If provenance is useful to describe a domain, topic and genre are also likely to be useful, and unlike provenance can often be identified given only the text of a sentence.
    \item While language domains certainly differ, they are not necessarily distinct: various topics and genres of language may overlap, while provenance is a discrete label.
    \end{itemize}  
    
Existing work on domain-specific translation tends to treat the domain of `test' language as known and/or completely distinct from other domains \shortcite{chu2018survey}. Known-domain translation may be relevant in limited scenarios, such as WMT shared tasks  where the topic or genre of text is pre-specified, or  bespoke translation systems adapted to customer data. Effective techniques for this  scenario include use of domain-specific neural network layers which are adapted while original parameters are frozen \shortcite{bapna-firat-2019-simple}.  Similar approaches have recently become popular throughout  natural language processing (NLP) for large pre-trained language models: BERT \shortcite{devlin-etal-2019-bert} and GPT-3 \shortcite{brown2020language} among others. These  models require a huge amount of pre-training data and computational investment, but once trained a small set of parameters can be tuned on a small dataset for good performance on a fixed task with relatively low cost.

However, even in fixed-task settings, knowledge of domain or provenance may not always help. A corpus domain label  may not be representative of all sentences in that corpus, and a sentence might reasonably be said to belong to more than one domain. In a very general case,  sentences supplied to a free online translation system could come from any source and contain features of any domain.  In this article we therefore explore not only work on improving translation of one  domain, but research into systems that can translate text from multiple domains: \emph{multi-domain} translation.  We will also emphasise techniques for  adaptation that can incorporate the benefits of a new domain without succumbing to forgetting, brittleness, overfitting, or  general failure to successfully translate anything other than a chosen set of adaptation sentences.

The survey will begin in section \ref{secdomain} by stating what we mean by a domain or multi-domain adaptation problem. Section \ref{sec:review-nmt} gives a brief overview of NMT to provide context for later adaptation approaches. In section \ref{sec:defaultdomainadaptation} we summarize fine-tuning, often used as a default or baseline approach to adaptation, and highlight  difficulties that motivate alternative approaches. We then review ways to control or improve domain adaptation at various stages in NMT system design, training and use. These can be data-centric (section \ref{litreview-dataselection}), change the neural network architecture (section \ref{litreview-architecture}) or adaptation procedure (section \ref{litreview-domainforgetting}), or take place at inference (section \ref{sec:review-inference}). We will explore the possibilities and pitfalls present at each step of system development.  Section \ref{litreview-casestudy} briefly describes three challenges for NMT that have  benefited from framing as domain adaptation problems: low-resource language translation,  demographic bias in translation, and document-level translation. Section \ref{sec:conclusions} concludes with a view of some open challenges in domain adaptation for NMT.

The structure of this survey aims to account for the fact that researchers and practitioners often only have control over certain elements of NMT system production. A typical scenario is commercial NMT providers allowing customers access to a pre-trained model with a fixed inference procedure which the providers adapt to customer data \shortcite{savenkov2018adaptive}. In this case, only the data used for adaptation can be affected by the user. By contrast, a researcher participating in a shared task might have a fixed set of adaptation data to make best use of, and  focus on adjusting the  network architecture or adaptation procedure itself. Even a user with access only to a black-box commercial tool may be able to customize the output with pre- and post-processing terminology control.

\citeA{chu2018survey} previously performed a brief survey on domain adaptation for NMT, later extended with some additional citations  \cite{chu2020survey}. 
While their survey focuses on a handful of domain adaptation methods, primarily mixed fine-tuning, instance weighting and architectural domain control, this article offers a  more substantive overview of adaptation techniques. 

In contrast to prior work we also emphasise the relative advantages and disadvantages of different methods in specific scenarios,  describe multi-domain adaptation, and discuss how the described techniques can be applied to NMT challenges not typically thought of as domains.

Specific contributions of this article are as follows:
\begin{itemize}
    \item We provide a thorough and up-to-date survey of domain adaptation techniques in popular use for machine translation at time of writing. 
    \item We provide a taxonomy of adaptation approaches corresponding to the process of developing an NMT system, letting practitioners identify the most relevant techniques. 
    \item We distinguish between domain adaptation and multi-domain adaptation, and describe relevant techniques for each.
    \item We focus throughout on various pitfalls in adaptation as they apply to the surveyed techniques, particularly forgetting and overfitting. 
    \item We explore the potential of framing other NMT research areas as adaptation problems.
\end{itemize}

\section{What Constitutes a Domain Adaptation Problem?}
\label{secdomain}

Adapting to a `domain' for machine translation has come to refer to a number of disparate concepts.  \shortciteA{van2017inadomain}  distinguishes various aspects that combine to form a domain for statistical MT. In this section, we briefly review their findings and the surrounding literature. We then clarify what is meant by a domain, and by a domain adaptation problem, for the purposes of this survey.

\subsection{Exploring the Definition of Domain}
\label{litreview-domaindescription}

Many possible categories and sub-categories may describe a language domain for purposes as disparate as education, text classification or data retrieval. These may not be well defined, especially across fields of research \shortcite{sinclair1996preliminary}. However, we concentrate on three primary elements of a domain  identified by \shortciteA{van2017inadomain} in the context of machine translation research: provenance, topic and genre.

\emph{Provenance} is the source of the text, usually a single discrete label. This may be a narrow description, such as a news article by a single author, or an extremely broad description, such as the 37M English-German web-crawled sentence pairs in the cleaned Paracrawl corpus \shortcite{banon-etal-2020-paracrawl}. Importantly, the provenance of a test sentence may not be known outside well-defined tasks in the research community, such as the WMT shared tasks.  Even if the provenance is known, it may not be useful for adapting translation performance unless it also corresponds to topic or genre. To take an  example from commercial MT, it may not be helpful to know that a sentence is generated by a particular customer unless their writing has specific features that should be reflected in a translation.  However, it is possible, if less common, to require domain-specific translations of sentences of a given provenance instead of using sentence content to determine domain. For example, we may have very similar or identical source sentences from two different customers who have different preferred translations for certain terms.

\emph{Topic} is the subject of text, for example news, software, biomedical. Topic often reveals itself in terms of distribution over  vocabulary items. Each word in the vocabulary may have different topic-conditional probabilities, and a document (or sentence) may be classified as a mixture of topics \shortcite{blei2003latent}. The topic(s) of a given sentence can also be determined as a mix of latent topics determined over a large dataset. Explicit or implicit topic may be used to resolve lexical ambiguity for translation. 

\emph{Genre} may be interpreted as a concept that is orthogonal to topic, consisting of function, register, syntax and style \shortcite{santini2004state}. For example, multiple documents about a company may share topics and use similar vocabulary, such as the company name or specific products. However, a recruitment document, product specification, or product advertisement would all constitute different genres \shortcite{lee2002text}. In some languages it is possible to convey  the same information about the same topic in multiple genres, for example by changing formality register. Various cues exist for automatic genre detection in text, including relative frequency of different syntactic categories, use of particular characters such as punctuation marks, and sentence length distribution \shortcite{kessler-etal-1997-automatic}.


\subsection{Domain Adaptation and Multi-Domain Adaptation}
\label{litreview-domainproblemdescription}

In this article, we identify a domain adaptation problem  as occurring  within the following broad circumstances:
\begin{enumerate}
    \item We wish to \emph{improve translation performance} on some set of sentences with identifiable characteristics. The characteristic may be a distinct vocabulary distribution, a stylometric feature such as sentence length distribution, some other feature of language, or otherwise meta-information such as provenance.
    \item We wish to \emph{avoid retraining} the system from scratch. Retraining is generally not desirable since it can take a great deal of time and computation, with correspondingly high financial and energy costs. Retraining may also be impossible, practically speaking, if we do not have access to the original training data or computational resources.
\end{enumerate}

We are also interested in \emph{multi-domain} adaptation. In  broad terms, a multi-domain adaptation problem occurs if, in addition to the previous two circumstances:

\begin{enumerate}
    \item[3.] We wish to achieve good translation performance on text from \emph{more than one domain}  using the same system. 
\end{enumerate}
We find the multi-domain scenario interpreted in a number of ways in the NMT literature. Some common settings for multi-domain NMT are as follows:

\begin{itemize}
    \item Domain labels are known for any given sentence. This is often assumed when adapting to multiple domains \shortcite{pham-etal-2021-revisiting}.  In this case multi-domain control or domain-specific subnetworks can be used, as described in section \ref{litreview-architecture}.
    \item Alternatively, domain membership for some sentences may be unknown, in which case it must be inferred. This is the case in the somewhat rarer setting where the domains of interest are considered to be fuzzy and potentially overlapping. The closest previously-seen domain can be inferred by a domain classifier subnetwork  as described in section \ref{subnetwork}, or accounted for at inference time as in section \ref{sec:review-inference}.
    \item Text from a labeled but unseen domain may appear at inference time. The system may classify it as the closest `seen' domain, as for unknown-domain text in the  previous point. Alternatively it may be possible to explore few-shot unsupervised adaptation, as described in section \ref{sec:syntheticdata}, or even zero-shot adaptation using priming techniques as in section \ref{sec:prepostprocess}.

    \item The domain of the pre-training data is one of the multiple domains we wish to be able to translate well. This is sometimes referred to as `continual learning' \shortcite{cao-etal-2021-continual}. Maintaining performance on the pre-training domain involves avoiding catastrophic forgetting during adaptation, a problem we discuss in section \ref{sec:forgetting}. While there are many potential solutions to this problem, most popular approaches involve adjusting training algorithm, discussed in section \ref{litreview-domainforgetting}.
\end{itemize}
Clearly it is possible for some of these multi-domain settings to co-occur. For example, we may want to avoid catastrophic forgetting on the original domain and also account for new domains at inference time. We may have domain labels during training, but need to infer domain at inference time. However, they are simpler to consider in isolation, and solutions addressing each individually can also often by combined.

In section \ref{litreview-casestudy} we  provide case studies on separate lines of machine translation research: low-resource language translation, gender handling in translation, and document-level translation. These  are well-established NMT research topics in their own right. Elements of at least the latter two may be viewed as relevant to genre or topic in terms of gender- and document-consistent grammar and lexical choice. However, they are not always treated as relevant to domain adaptation. We will demonstrate that these lines of translation research can also be treated as domain adaptation problems.

\section{A Brief Overview of Neural Machine Translation}
\label{sec:review-nmt}
Machine Translation (MT) aims to translate written text from a source natural language to a target language. Originally accomplished with statistical MT (SMT) using phrase-based frequency models, the state-of-the-art  in high-resource language pairs like English-German has been achieved by Neural Machine Translation (NMT) since 2016 \shortcite{bojar-etal-2016-findings}. More recently NMT has also outperformed SMT on low-resource language pairs  on generic domains \shortcite{sennrich-zhang-2019-revisiting} or those with sufficient domain-specific data  \shortcite{DBLP:conf/euspn/AhmadniaD20}. Here we summarize NMT,  focusing on elements relevant to the domain adaptation techniques in this survey. For an in-depth discussion of general approaches to NMT, we direct the reader to broader surveys, e.g. \shortciteA{koehn2020neural},  \shortciteA{stahlberg2020neural}. 

\subsection{Representing Language for Neural Machine Translation}
\label{litreview-datarepresentation}
NMT models usually take as a single training example two token sequences: $\mathbf{x}$ representing a source language sentence and $\mathbf{y}$ representing its target language translation. Tokens are represented as integer IDs $\in V$ where $V$ is the source or target vocabulary, which may represent words, subwords, or other features such as domain tags.  $|V|$  is usually limited to tens of thousands of tokens due to the computational complexity of the softmax used to map  embeddings to discrete tokens.  The vocabulary should convey information that is useful for translation, while remaining computationally tractable. Unknown or out-of-vocabulary (OOV) tokens are undesirable, as are large vocabulary sizes and very long sentence representations. This causes trade-offs -- for example, a character vocabulary might be very small with no OOVs but very long sequences.  Here we describe ways to address these trade-offs when representing language for NMT. The chosen approach often affects domain adaptation, since vocabulary and terminology are strongly indicative of domain.

\subsubsection{Word Vocabularies}
\label{litreview-datarepresentation-word}
Early approaches to NMT focused on word vocabularies, usually either the top $|V|$  words by frequency \shortcite{cho-etal-2014-learning} or all training words with frequencies over a threshold \shortcite{kalchbrenner-blunsom-2013-recurrent-continuous}. Out-of-vocabulary (OOV) words are represented by a special \texttt{UNK} token. Word vocabularies must account for sparsity. According to \shortciteA{zipf1949human} a word's occurrence probability is inversely related to its vocabulary rank: a large proportion of  words in a language are rare in a given corpus. Representing every  rare word as a unique vocabulary item is inefficient, and may not allow good learned representations.  



\subsubsection{Subword Vocabularies}
\label{litreview-datarepresentation-bpe}

Subword vocabularies address the rare word problem by representing words as sequences of more frequent subwords. \shortciteA{sennrich2016subword} first propose NMT on subword sequences  using  Byte Pair Encoding (BPE)  \shortcite{gage1994new}. A BPE vocabulary initializes with the set of all character tokens, and with all  words represented as character sequences. BPE then iteratively merges the most frequent token pair. Frequent words  become single tokens, while  rare words can be represented as character sequences. The most extreme subword segmentation is a character vocabulary \shortcite{ling2015character,kim2016character,costa-jussa-fonollosa-2016-character,cherry-etal-2018-revisiting}. Subword vocabularies have been proposed using  syllables \shortcite{assylbekov-etal-2017-syllable}, language model scores \shortcite{kudo-2018-subword} or linguistically-informed  subwords \shortcite{ataman2017linguistically,huck-etal-2017-target,machavcek2018morphological}.  However, BPE  is currently widely accepted as a default vocabulary scheme for NMT. Much recent work explores optimizing BPE granularity. \shortciteA{ding-etal-2019-call} find that low-resource NMT benefits from fewer BPE merges  to maintain frequency for individual subwords. \shortciteA{galle-2019-investigating} and \shortciteA{salesky2020optimizing} similarly find best BPE performance when balancing high-frequency subwords with short sequence lengths overall. 


\subsubsection{Tags}
NMT language representations primarily vary in granularity, from words to characters. Alternatively, sentence representations can be augmented with  elements not present in the original text. One example  relevant to domain adaptation uses externally-defined tags to indicate a particular feature of the sentence. Previous work has used sentence tags to convey domain \shortcite{kobus2017domain,britz-etal-2017-effective}, speaker gender \shortcite{vanmassenhove-etal-2018-getting}, or target language formality \shortcite{sennrich-etal-2016-controlling,feely-etal-2019-controlling}. Multiple tags can be used throughout source and target sentences, for example indicating linguistic features \shortcite{sennrich-haddow-2016-linguistic,garciamartinez2016factored,aharoni2017towards,saunders-etal-2018-multi}
 or  custom terminology use \shortcite{dinu-etal-2019-training,michon-etal-2020-integrating}. Tags are usually incorporated throughout training, implicitly requiring the availability of reliable tags for large training datasets, although new tags can also be introduced at adaptation time \shortcite{saunders2020neural}. Tags can  be treated as any other token in a sentence, or handled in special ways by the neural network (section \ref{litreview-architecture}). 


\subsubsection{Representing Extra-Sentence Context}
\label{sec:doc}
NMT typically translates between sentence pairs, but prior work has incorporated source context, from the previous  sentence \shortcite{tiedemann2017neural} to the whole document \shortcite{junczys-dowmunt-2019-microsoft,mace2019using}. Early work on context for recurrent NMT showed improvements in BLEU score \shortcite{wang-etal-2017-exploiting-cross} and context-based metrics like pronoun accuracy \shortcite{jean2017does}. For self-attention models, context improves  lexical cohesion and anaphora resolution \shortcite{voita-etal-2019-good,voita-etal-2018-context}, but has inconsistent effects on overall quality  \shortcite{stahlberg-etal-2019-cued}.  Notably, \shortciteA{kim-etal-2019-document} suggest that context use may simply regularize NMT, and that context-specific information for e.g. lexical cohesion can be retained in very minimal forms, such as  terminology tags or syntactic annotation -- much like domain tags. While we focus on sentence-level NMT in this article,  in section \ref{sec:docnmt} we describe  approaches to document-level NMT that take a domain adaptation approach.

\subsection{Neural Translation Model Architecture}
\label{litreview-nmtarchitecture}

To the NMT architecture, all inputs are sequences of generic `tokens', represented as integer values. The NMT model input is  $\mathbf{x} = x_1, ..., x_I, x_i \in V_{src}$, and  it must produce another integer sequence: $\mathbf{y} = y_1, ..., y_J, y_j \in V_{trg}$. We refer to the phase of updating model parameters given $\mathbf{x}$ and $\mathbf{y}$ as training, to the phase where parameters are fixed and we generate hypotheses $\mathbf{\hat{y}}$ without reference tokens $\mathbf{y}$ as inference, and to the process of producing any sequence with the decoder as decoding. While many NMT networks can learn mappings between $\mathbf{x}$ and $\mathbf{y}$ that generalize to unseen $\mathbf{x}$ during inference,  we focus on the Transformer  \shortcite{vaswani2017attention}, the de facto standard for NMT in recent years.

\subsubsection{Continuous Token Embeddings}
\label{litrevew-wordembed}
A Transformer first maps a sequence of integers  $\mathbf{x} \in V_{src}$ to  embeddings.  A $|V|>10$K-dimension token representation is intractable when operating on thousands of tokens per batch. It is also discrete: the effect of changing one token may not generalize. \shortciteA{bengio2003neural} proposed instead mapping each vocabulary item to a continuous feature vector or `embedding'. These are a more tractable size: typically $d\leq1024$.  
Embeddings trained with context-related objectives -- as in NMT -- may exhibit local smoothness and therefore generalize: words with similar contexts tend to have similar embeddings  \shortcite{collobertweston-08-embeddings,turian-etal-2010-word}.  In particular embeddings are often similar for tokens that belong to the same category  -- that is, the same domain \shortcite{collobert-2011-nlp,mikolov2013distributed}. 
These properties can aid data selection for adaptation (section \ref{litreview-dataselection}). 

\subsubsection{Transformers: Attention-Based Encoder-Decoder Networks}
\label{litreview-nmtarchitecture-transformer}

\begin{figure}[ht]
\begin{center}
\includegraphics[scale=0.4]{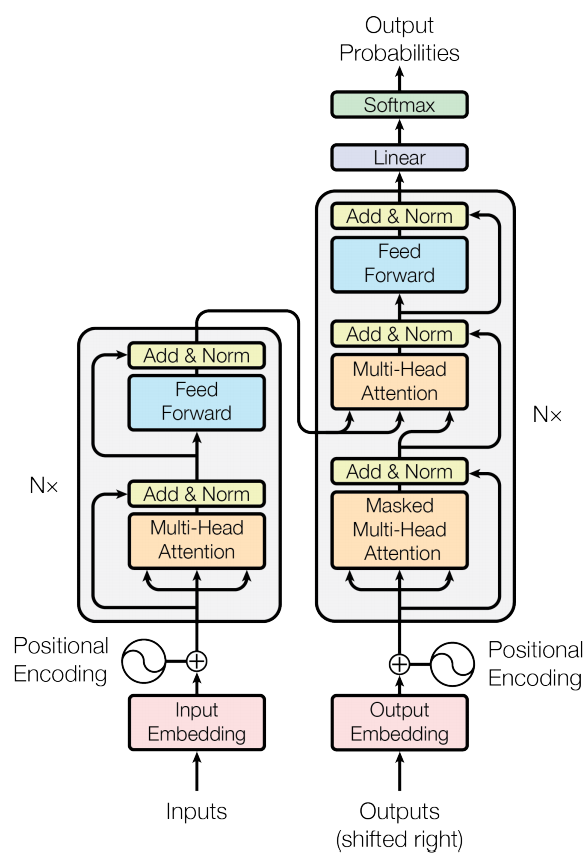}
\end{center}
 \caption{Illustration of the Transformer model architecture, Figure 1 from \shortciteA{vaswani2017attention}. The model is made up of encoder (left) and decoder (right) subnetworks.}
\label{fig:transformer}
\end{figure}

The Transformer network, proposed by \shortciteA{vaswani2017attention} and illustrated in Figure \ref{fig:transformer}, is a self-attention-based encoder-decoder model. It remains state-of-the-art for NMT at time of writing  \shortcite{barrault-etal-2020-findings}, improving parallelizability and quality over earlier RNN-based NMT models. The Transformer encoder  produces a source sentence embedding from the source token embeddings. Its decoder produces a target  embedding from the target tokens during training or the partial translation hypothesis at inference.  An encoder-decoder attention network relates the encoder output and decoder state. At inference the decoder translates based on the target embedding and the encoder-decoder attention. 

Recurrent model attention mechanisms such as the earlier state-of-the-art described in \shortciteA{bahdanau15jointly} related different positions in different sequences in order to learn a `context representation'.  The Transformer's self-attention embedding for a sequence is calculated from  different positions projected from the same sequence. Source embeddings $x$ are used for self-attention at the encoder input and target embeddings $y$ at the decoder input. For a multi-layer encoder or decoder self-attention is calculated on the output of the previous layer. Any of these elements - encoder, decoder, and attention network - can be made domain-specific, or duplicated for a specific domain (section \ref{subnetwork}). 


\subsubsection{Increasing Model Depth}

 \label{litreview-multilayer}
The Transformer encoder and decoder are usually formed of several identical layers, each  operating on the output of the layer before. A deeper model can experience training difficulties, as loss gradients must propagate through more layers. Difficulties can be mitigated by adding residual networks, changing each layer output $f(z)$ to $f(z) + z$ \shortcite{he2016deep}. This gives each encoder or decoder layer  access to the first layer's input. Residual networks are also needed for some domain-specific subnetworks like adapter layers \shortcite{bapna-firat-2019-simple} (section \ref{sec:adapter}). Deep models can significantly improve NMT \shortcite{wang-etal-2019-learning}, but shallower models still often perform better for low-resource NMT \shortcite{sennrich-zhang-2019-revisiting,nguyen-chiang-2018-improving}.  The  potential pitfalls of deeper models are worth considering in relation to architectural approaches to domain adaptation (section \ref{litreview-architecture}).


\subsection{Training Neural Machine Translation Models}
\label{litreview-nmttraining}
Once an NMT model architecture is determined its parameters must be adjusted to map from source  $\mathbf{x}$ to target $\mathbf{y}$. NMT model parameters are trained by backpropagation \shortcite{rumelhart1986learning}, usually with a Stochastic Gradient Descent (SGD) optimizer, which requires some objective on the training set. Standard training objectives such as cross-entropy loss use real reference sentences, but during inference only the  prefix of the model's own hypothesis $\hat{\mathbf{y}}$ is available.  This difference in conditioning can cause poor behaviour during inference \shortcite{bengio2015scheduled,ranzanto16sequencelevel}. Avoiding this over-exposure to training examples has motivated parameter and objective regularization methods during training. Here we focus on  objective function variations which are relevant to domain adaptation.

\subsubsection{Cross-Entropy Loss}
\label{litreview-crossentropy}
The most common NMT training objective is varying weights $\theta$ in the gradient direction of  log likelihood of training examples to maximize log likelihood. This is known as Maximum Likelihood Estimation (MLE) \shortcite{baum1988supervised,levin1988accelerated}.

\begin{equation}
\hat{\theta} = \argmax_{\theta}\log P(\mathbf{y}|\mathbf{x}; \theta)
\end{equation}

MLE is equivalent to minimizing cross-entropy loss $L_{CE}$ between the generated output distribution and the references if each token has one reference label $q(y'=y_j|\mathbf{x};\theta) = \delta(y_j)$:

\begin{equation}
\hat{\theta} = \argmin_{\theta}\sum_{j=1}^{|y|}-\log P(y_j|\hist, \mathbf{x}; \theta) = \argmin_{\theta}L_{CE}(\mathbf{x}, \mathbf{y}; \theta)
\label{eq:minceloss}
\end{equation}

This survey will describe several variations on the MLE loss for domain adaptation (section \ref{litreview-domainregularization}), as well as adaptation-specific applications of non-MLE objectives which seek to address some of the downsides of MLE (section \ref{litreview-mrt}).

\subsubsection{Objective Function Regularization} 
We can  change Eq. \ref{eq:minceloss} by adding a regularization term $L_{\text{Reg}}$  to the loss function itself:
  
\begin{equation}
\hat{\theta} = \argmin_{\theta} [L_{CE}(\mathbf{x}, \mathbf{y}; \theta) + \lambda L_{\text{Reg}}(\theta) ]
\label{eq:multitaskloss}
\end{equation}

One option is an L2 penalty term, $L_{\text{Reg}} = \sum_i\theta_i^2$. $L_{\text{Reg}}$ can also be an objective from another task, known as multi-task learning.  Translation-specific multi-task terms include a coverage term \shortcite{tu-etal-2016-modeling}, a right-to-left translation objective \shortcite{zhang2019regularizing}, the `future cost' of a partial translation \shortcite{duan2020modeling}, or a target language modelling objective \shortcite{gulccehre2015using,sriram2018cold,stahlberg-etal-2018-simple}.  Another approach is dropout: randomly omitting a subset of parameters $\theta_{\text{dropout}}$ from optimization for a training batch \shortcite{hinton2012improving}. This corresponds to  regularization  with $ L_{\text{Reg}}(\theta)=\infty$ for $\theta \in \theta_{\text{dropout}}$,  0 otherwise.  Objective function regularization is often used to avoid catastrophic forgetting during NMT domain adaptation. We discuss methods relating to dropout in section \ref{sec:freeze}, and other objective regularization methods in section \ref{sec:regularize}.

\subsubsection{Output Distribution Regularization}
\label{sec:distribreg}
MLE assumes that a reference sentence is far more likely than any other translation. This encourages large differences in likelihood between training examples and language not seen in training: overfitting. Overfitting can reduce the model's ability to cope with novel data during inference. This is especially relevant when adaptation to small domains with limited training examples. Overfitting can be mitigated by regularizing the distribution over output labels during training, for example using label smoothing \shortciteA{szegedy2016rethinking}.
This replaces the single target label  $q(y'|\mathbf{x};\theta) = \delta(y_j)$ used to derive Eq. \ref{eq:minceloss}, instead smoothing the label distribution by hyperparameter $\epsilon$ towards a uniform distribution over the vocabulary:

\begin{equation}
 q(y'|\mathbf{x};\theta) = (1 - \epsilon) \delta(y_j) + \frac{\epsilon}{|V_{trg}|} 
\end{equation}

\begin{equation}
\hat{\theta} = \argmin_{\theta}\sum_{j=1}^{|y|}\sum_{y' \in V_{trg}} - q(y'|\mathbf{x};\theta) \log P(y'|\hist, \mathbf{x}; \theta) 
\label{eq:mincelsloss}
\end{equation}

Instead of smoothing with a uniform
distribution $\frac{1}{|V_{trg}|}$, label smoothing can incorporate prior information about the target language. For example, the smoothing distribution can come from a larger `teacher' model \shortcite{Hinton2015DistillingTK}. \shortciteA{pereyra2017regularizing} explore  smoothing towards a unigram distribution over the vocabulary, and addressing over-confidence directly by penalizing peaky, low-entropy output distributions. Related distribution regularization schemes  have been applied to domain adaptation (section \ref{sec:KD}).


\subsection{Inference with Neural Machine Translation Models}
\label{litreview-nmtinference}
An NMT model trains on  source and target language sentences $\mathbf{x}$ and $\mathbf{y}$, and learns to model $P(\mathbf{y}|\mathbf{x})$. At inference, the model has access only to $\mathbf{x}$, and must produce a translation:

\begin{equation}
\hat{\mathbf{y}} = \argmax_{\mathbf{y}} P(\mathbf{y}|\mathbf{x})
\end{equation}

In this survey we focus on autoregressive NMT inference, by far the most common approach. For autoregressive inference the model produces one output token at each inference step $j$, which is  conditioned on the source sentence and all previously output tokens. Ideally: 
\begin{equation}
\hat{y_j} = \argmax_{y_j} P(y_j|\hist, \mathbf{x})\label{eq:l2rinference}
\end{equation}
$|V_{trg}|^j$ possible partial translations end in the $j^{th}$ output token. Exploring all of these is impractically slow even using likelihood pruning to reduce the total \shortcite{stahlberg-byrne-2019-nmt}. Nevertheless, approximations to this inference objective work well in practice. Here we focus on inference procedures which have variants used for domain-specific translation.

\subsubsection{Beam Search}
\label{litreview-beamsearch}
Beam search is the most common approximation for NMT inference. 
It tracks the top $N$ partial hypothesis `beams' by log likelihood. At each inference step all possible single-token expansions of all beams are ranked, and the new top $N$ selected. Search continues until all beams terminate with an end-of-sentence token or exceed a maximum length. `Greedy' search is the special case where $N=1$, which produces the most likely next token at each step. Beam search variations include optimizing for diverse hypotheses \shortcite{vijayakumar2016diverse,li2016mutual} or  adequacy \shortcite{wu2016google}. In section \ref{sec:domainrescore} we describe beam search variants intended to produce domain-specific translations.

\subsubsection{Ensembling}
\label{litreview-ensembling}
Conducting inference with an ensemble of MT models allows consensus on which tokens to expand and track in beams at each inference step, usually outperforming individual models \shortcite{sim2007consensus,rosti-etal-2007-combining}. Ensembles can integrate scores from different NMT architectures \shortcite{stahlberg-etal-2018-university} or language models \shortcite{vaswani-etal-2013-decoding,gulccehre2015using}.  Many schemes for ensembling exist, from majority vote to minimizing Bayes risk under some metric \shortcite{rokach2010ensemble}. Ensembling by static weighting is commonly used in NMT. For a $K$-model ensemble with a weight $W_k$ for each model, the ensemble translates: 
 
\begin{equation}
\hat{\mathbf{y}} =  \argmax_\mathbf{y} P(\mathbf{y} |\mathbf{x}) = \argmax_\mathbf{y}  \sum_{k=1}^K W_k  P_k(\mathbf{y} | \mathbf{x})
\label{eq:weight-ensemble}\end{equation}


A downside of ensembling is reduced efficiency. The expensive softmax calculation takes place $K$ times, and often all ensemble models are stored in memory simultaneously. Efficiency can be improved if one model can learn to reproduce the ensemble's behaviour, using simplification schemes like ensemble knowledge distillation \shortcite{freitag2017ensemble,Fukuda2017EfficientKD} or ensemble unfolding \shortcite{stahlberg-byrne-2017-unfolding}. Many approaches to ensembling or ensemble simplification assume that all ensembled models generate in the same way, e.g. lexical items from the same vocabulary, and that the same weighting is appropriate for all input sentences regardless of text or model domain. In section \ref{sec:review-multidomainensemble} we review domain-specific ensembling  approaches which relax these assumptions.



\section{Fine-Tuning as a Domain Adaptation Baseline, and its Difficulties}
\label{sec:defaultdomainadaptation}
While the previous section covered `default' approaches to  NMT, e.g. a Transformer with a  BPE vocabulary trained with cross-entropy loss, this section covers fine-tuning, a `default' approach to domain adaptation for NMT. It  also describes some difficulties with this approach which motivate many variations  in the remainder of this survey.

Given an in-domain dataset and a pre-trained neural model, domain adaptation can often be achieved by continuing training --   `fine-tuning' -- the model on that dataset. Fine-tuning initially involved monolingual data only for SMT \shortcite{lavergne2011iwslt}. By contrast fine-tuning end-to-end NMT models, to the best of our knowledge first proposed in \shortciteA{luong2015stanford}, usually requires bilingual parallel data. In both cases the pre-trained model trains for relatively few additional iterations with the original loss function now applied to the in-domain data. This is  a straightforward and  efficient approach: adaptation sets might have   thousands of sentence pairs, while training the original model could use millions.  \shortciteA{luong2015stanford} find that in-domain NMT can improve dramatically under fine-tuning despite the low computational requirements. \shortciteA{senellart2016domain} observe that fine-tuning on in-domain data may not reach the improvements possible when retraining the entire system on combined generic and in-domain data, but report a high proportion of the improvement in less than one percent of the time needed for complete retraining.

Possible variations on fine-tuning for domain adaptation include changing the data, changing the network architecture, or changing the tuning procedure itself. Alternatively, domain-specificity can instead by achieved with  domain-adaptive inference. Multiple of these variations on simple fine-tuning  can be applied in combination. For example, a variation on fine-tuning could  tune on a synthetically expanded in-domain set, using domain-specific subnetworks, with objective regularization to avoid forgetting or overfitting. However, we note that in the literature it is common to compare proposed domain adaptation interventions in isolation to a baseline of simple fine-tuning on a pre-defined dataset.

Simple fine-tuning is associated with a number of potential difficulties. We present three here -- insufficient tuning data, forgetting, and overfitting --  as motivation for the variations on fine-tuning data, architecture, tuning procedure and inference procedure that are presented in the remainder of this survey.

\subsection{Difficulty: Not Enough In-Domain Data}
Fine-tuning strong pre-trained models on small, trusted sets has become a popular approach for machine translation fixed-domain  tasks. A version of this commonly appearing in WMT shared task submissions is tuning on test sets released for the same  task in previous years  \shortcite{schamper-etal-2018-rwth,koehn-etal-2018-jhu,stahlberg-etal-2019-cued}. Shared task adaptation sets are likely to contain many sentences stylistically very similar to those in the test set. However, it is not always possible to assume the existence of a sufficiently large and high-quality in-domain parallel dataset. Outside  well-defined domain adaptation shared tasks, there may be little or no bilingual in-domain data in the required languages available for a particular adaptation scenario. In section \ref{litreview-dataselection}, we describe data-centric adaptation methods which aim to expand the  in-domain parallel corpus available for tuning.

\subsection{Difficulty: Forgetting Previous Domains}
\label{sec:forgetting}
The difficulty of `catastrophic forgetting' occurs when translating domains other than the domain of interest. If a neural model with strong performance on domain $A$ is fine-tuned on domain $B$, it often gains strong translation performance on domain $B$ at the expense of extreme performance degradation on domain $A$ \shortcite{mccloskey1989catastrophic,ratcliff1990connectionist}. This is especially problematic if the domain of the test data may not be the last domain the model tuned on. For multi-domain adaptation, we may want a single model that is capable of translating text from multiple domains, regardless of when during training a given domain was learned.

Some forgetting may be permissible if the model is intended to translate only a small amount of highly specific data. Examples include adapting a new model to translate each individual test sentence \shortcite{farajian-etal-2017-multi,li-etal-2018-one,mueller-lal-2019-sentence} or document \shortcite{kothur-etal-2018-document}. However, we may want to maintain good performance on previously seen domains. Even in this scenario, forgetting  all data that is unlike the tuning data may cause poor performance on even very specific target-domain data -- this is the highly related scenario of overfitting (section \ref{ref:overfit}). 

A straightforward approach to avoiding forgetting under fine-tuning is to simply tune for fewer steps \shortcite{xu2019lexical}, although this introduces an inherent trade-off between better performance on the new domain and worse performance on the old domain. Other approaches to good  performance on both new and old domains with a single model might introduce additional domain-specific parameters or subnetworks for the new domain, as described in section \ref{litreview-architecture}. Alternatively they may involve changing the training approach, for example to add regularization relative to the original domain, as described in section \ref{litreview-domainforgetting}.


\subsection{Difficulty: Overfitting to New Domains}
\label{ref:overfit}
Overfitting or `exposure bias' is common when the fine-tuning dataset is very small or repetitive. The model may be capable of achieving excellent performance on the precise adaptation domain $B$, but any small variations from it -- say, $B'$ -- cause difficulties. \shortciteA{farajian-etal-2017-multi} and \shortciteA{li-etal-2018-one} observe this degradation when adapting a model per-sentence on only tens of training sentences that are not sufficiently similar to the sentence of interest if the learning rate or number of adaptation epochs is too high. Overfitting is particularly relevant in cases of domain mismatch between the test sentence domain and the fine-tuning domain \shortcite{wang-sennrich-2020-exposure}.  

The effect can also occur when the tuning data contains small irregularities. These may be idiosyncratic vocabulary use, or misaligned sentence pairs where source and target sentences are not good translations. In this case test sentences may  match the tuning data's topic and genre, but we   wish them to fall into $B'$  -- domain where target sentences are good translations of the source -- rather than $B$ -- domain with irregularities. \shortciteA{DBLP:conf/icml/OttAGR18} observe that a model trained on data with too high a noise proportion may fall back on language model behaviour and generate some common target sentence from the  dataset, or even simply copy the input sentence. \shortciteA{saunders-etal-2020-exposure} observe that even when adapting to tens of thousands of sentence pairs, tens or hundreds of systematically misaligned sentence fragments in the adaptation set can trigger undesirable nonsense translations after tuning.    

Overfitting can be mitigated  by expanding the in-domain corpus so that it is harder to overfit (section \ref{litreview-dataselection}), or by adjusting the adaptation procedure to adapt less aggressively to the new domain (section \ref{litreview-domainforgetting}). An alternative which sidesteps fine-tuning, and therefore much of the potential for overfitting or forgetting, is adjusting the test-time inference procedure without changing the model at all (section \ref{sec:review-inference}).

\section{Data-Centric Adaptation Methods}
\label{litreview-dataselection}
A domain is often identifiable by features of its data. Topic and genre as described in section \ref{litreview-domaindescription} are defined in terms of vocabulary and syntactic style for in-domain text. However, the existence of a sufficiently large in-domain parallel dataset is often taken for granted by fine-tuning (section \ref{sec:defaultdomainadaptation}), as well as many model adaptation methods described in sections \ref{litreview-architecture} and \ref{litreview-domainforgetting} of this survey.  Here we consider how an in-domain dataset can  be created or expanded.

By data-centric adaptation methods, we refer to methods revolving around selecting or generating appropriate in-domain data. In the absence of a pre-defined in-domain corpus, natural data  -- produced by a human -- may be selected from a larger generic corpus or corpora according to some domain-specific criteria.  A special case of natural data selection is filtering an existing in-domain corpus for cleaner or more relevant data. 

Departing from exclusively natural data,   additional source or target sentences can be generated by  neural models if only monolingual natural data is available. This may be beneficial if the alternative is very sparse language or text that is mismatched with the domain of interest.
Semi-synthetic bilingual data can be produced by forward or back translation of existing natural data. Additional in-domain bilingual data can be partially synthesised from a small in-domain natural dataset using  noising or simplification. Finally, purely synthetic data can be synthesised for adaptation.


\subsection{Selecting Additional Natural Data for Adaptation}
\label{sec:selectdata}

Given a test domain of interest and a large pool of general-domain sentences, there are many ways to extract relevant data. We divide these into methods using discrete token-level measures for retrieval, those using continuous sentence representations for retrieval, or those that use some external model for scoring.

Sentences can be selected by discrete token content overlap with in-domain data. Many such methods were applied to statistical MT, which typically does not involve continuous sentence representations. For example, \shortciteA{eck-etal-2004-language} retrieve sentences for SMT language model adaptation by TF-IDF and n-gram-overlap relevance measures.  The NMT adaptation scenario where few in-domain sentences are available has likewise seen fine-tuning improvements when selecting additional data by straightforward n-gram matching, as demonstrated for single-sentence adaptation by \shortciteA{farajian-etal-2017-multi} and \shortciteA{li-etal-2018-one}. \shortciteA{koehn-senellart-2010-convergence} instead use a fuzzy match score based on word-level edit distance to retrieve similar source sentences for SMT.  \shortciteA{xu2019lexical} apply fuzzy matching retrieval to NMT,  using retrieved sentences as in-domain adaptation data for each test sentence. However, they obtain slightly better results using n-gram matching. Other applications of fuzzy matching for NMT use the retrieved sentences directly at inference time as for the \shortciteA{koehn-senellart-2010-convergence} approach -- we discuss these in section \ref{ref:prime}.

Another option is to select sentences that are similar to test inputs  but \emph{different} from previously selected adaptation sentences.  \shortciteA{poncelas-etal-2019-transductive} take this approach with the aim of maximising diversity in the adaptation data. They select either by overall n-gram rarity, or by using Feature Decay Algorithms which adjust the `value' of an n-gram based on how often it has been sampled. They find both approaches outperform TF-IDF similarity selection, which does not account for retrieved dataset diversity.

Tuning on data selected by discrete lexical overlap measures can be highly effective and also relatively efficient. For example, \shortciteA{farajian-etal-2018-eval} and \shortciteA{li-etal-2018-one} both note that adapting to data obtained by sentence-level  similarity with test sentences can noticeably improve  terminology use after adaptation, while requiring only a few adaptation examples per test sentence. However, fine-grained adaptation can result in overfitting if matching is imperfect, with \shortciteA{chen-etal-2020-character} finding better performance tuning on sentences selected by n-gram match over the entire test set compared to matching per-sentence.

In-domain data can also be retrieved based on continuous representations. For example, \shortciteA{wang2017sentence} select sentences with embeddings similar to in-domain sentence embeddings to add to the in-domain corpus. \shortciteA{bapna-firat-2019-non} combine  n-gram-based retrieval with
dense vector representations to improve candidate retrieval across multiple domains. Whether using discrete or continuous representation, matching usually occurs in the source language, as source sentences can be compared directly with test input sentences. Notable exceptions are the recent  approaches of \shortciteA{DBLP:conf/acl/00020LLL20} and \shortciteA{vu-etal-2021-generalised}, both of whom use a cross-lingual  network to retrieve target language sentences with similar embeddings to an input source sentence. While this  requires a strong cross-lingual neural model, it has the advantage that `real' relevant target sentences are retrieved. Many  previously-described approaches either rely on the retrieved source having a well-aligned target sentence, or retrieve only source sentences and then generating a synthetic target sentence (section \ref{sec:forwardtranslation}).

Sentences can be selected after explicit domain-relevance scoring by external models. \shortciteA{moore-lewis-2010-intelligent} select data for in-domain language model training by scoring the data under in-domain and general domain language models, and taking the cross-entropy difference, effectively relevance to the target domain. \shortciteA{axelrod-etal-2011-domain} add a bilingual cross-entropy difference term to the \shortciteA{moore-lewis-2010-intelligent} approach to select parallel data for SMT adaptation.  More recently, \shortciteA{DBLP:conf/ecir/VuM21} select documents based on support vector machines trained to classify between the target domain data and random mixed-domain data.   However, \shortciteA{van-der-wees-etal-2017-dynamic} note that such static discriminative filtering schemes can struggle with very similar  in-domain and generic corpora.   \shortciteA{axelrod2017cynical} instead suggests `cynical data selection': repeatedly selecting the sentence that most reduces the relative entropy for modelling the domain of interest.  Importantly given the ambiguity over the meaning of `domain' mentioned in section \ref{litreview-domaindescription}, this does not involve actually defining a domain. Indeed, \shortciteA{santamaria2017data}  note that using classifiers for data selection is not necessarily a good conceptual approach since a sentence pair may easily appear in multiple corpora, and instead reframe the `in-domain' and `general domain' corpora as data that we know we are interested in or do not yet have an opinion about. 

\shortciteA{aharoni-goldberg-2020-unsupervised} dispense with assigned corpus  labels and show that adapting to data identified by unsupervised domain clustering using large language models matches or out-performs tuning on the `correct' domain-labelled data. \shortciteA{DBLP:journals/corr/abs-2109-07864} extend this idea to show that  NMT models themselves can cluster sentences by embedding into domains that allow for better adaptation than pre-defined corpora. 
We regard these results as an indication that arbitrary labels should not be relied upon as domain identifiers.

\subsection{Filtering Existing Natural Data}
Use of provenance -- taking an existing  corpus label as indicative of domain -- is by far the simplest approach to natural data selection for adaptation, despite the deficiencies mentioned at the end of section \ref{sec:selectdata}. In the case where we do take a full corpus as in-domain, data filtering can be applied to ensure that the selected data is actually representative of a domain. For example, \shortciteA{taghipour2011parallel} map sentences in a parallel corpus to a feature space, and mark the most novel pairs in the feature space as noise to be removed. Care must be taken not to diminish the training space too far: for example, \shortciteA{lewis-eetemadi-2013-dramatically} attempt to maximize n-gram coverage with remaining data while filtering  sentences  for SMT.

A special case of data filtering is targeted to remove `noisy' training examples, as for example in the WMT parallel corpus filtering task \shortcite{koehn-etal-2018-findings}. Data cleaning may involve ensuring  source and target sentences in the training sentence are well-aligned, contain the languages of interest, are not too long or contain too many non-words, such as HTML tags \shortcite{khayrallah-koehn-2018-impact,berard-etal-2019-naver}. Cleaning methods can be very similar to domain data selection, for example with extensions of previously-mentioned cross-entropy difference filtering \shortcite{moore-lewis-2010-intelligent} to bilingual training examples, where the `in-domain' models are trained on verifiably clean data only \shortcite{junczys-dowmunt-2018-dual,junczys-dowmunt-2018-microsofts}.  While data cleaning has become a standard step in data preprocessing when training any NMT system, we note it is particularly crucial to domain adaptation if only a small in-domain dataset is available, as discussed in section \ref{ref:overfit}, and is widely mentioned in the submissions for domain-specific shared tasks \shortcite{bawden-etal-2019-findings}.

\subsection{Generating Synthetic Bilingual Adaptation Data from Monolingual Data} 
\label{sec:syntheticdata}

Bilingual training data that is relevant to the domain of interest may not be available. However, source or target language monolingual data in the domain of interest is often much easier to acquire. In-domain monolingual data can be used to construct partially synthetic bilingual training corpora by forward- or back translation. This is a case of bilingual data \emph{generation} for adaptation rather than data selection.

Source or target language sentences in the  domain of interest may be extracted from monolingual corpora using techniques described previously in this section. A sufficiently strong source-to-target translation model can forward-translate these sentences, or a target-to-source model can back-translate them. The result is aligned in-domain source and target language training sentences. This allows `unsupervised' MT domain adaptation where parallel data is unavailable. Statistical MT has benefited from both forward translation \shortcite{schwenk08investigations} and back-translation \shortcite{bertoldi-federico-2009-domain,lambert-etal-2011-investigations} in the context of domain adaptation, with back translation performing better in direct comparisons. Here we discuss more recent work on adaptation data generation for NMT.


\subsubsection{Back Translation}
\label{sec:backtranslation}
Back translation uses  natural monolingual data as target sentences, and requires a target-to-source NMT model to generate synthetic source sentences. 
Back translations are commonly used to augment general domain translation corpora, with strong improvements over models not trained on back-translated data \shortcite{sennrich2016improving}. Even models trained exclusively on back translations have shown similar  BLEU score to models trained on natural data \shortcite{poncelas2018investigating}. More relevant to this survey, back translation is also frequently used to produce data for adaptation  to a new domain where  monolingual in-domain target language sentences are available \shortcite{sennrich2017university,jin2020simple}. In-domain targets sentences may be identified with known topic and genre, for example biomedical papers  \shortcite{abdul-rauf-etal-2020-limsi} or  coronavirus-related texts  \shortcite{mandieh-rapid-2020}. \shortciteA{parthasarathy-etal-2020-adapt} extract target sentences by finding target language terms that are likely to be in the test set, while \shortciteA{vu-etal-2021-generalised} use a  cross-lingual domain classification model to extract target sentences  that are relevant to test source sentences. These target sentences are then back-translated.  The synthetic-source sentence pairs are typically used  directly for fine-tuning the model, but can also be used as candidates for a domain-specific data selection scheme \shortcite{poncelas-way-2019-selecting}.

Notably, \shortciteA{DBLP:journals/corr/abs-1906-07808} show that back-translated sentences can improve domain adaptation even when the original examples were used to train the general domain model, suggesting it is not simply the in-domain target sentences but the novel synthetic source sentences that aid adaptation.  However, \shortciteA{wei-etal-2020-iterative} argue that domain-relevance of synthetic sentences for domain adaptation  is particularly important, and that back-translating in-domain data using a general domain model can give poor synthetic source sentences due to the domain mismatch, harming adaptation.  They show improved adaptation by jointly learning a separate network to `domain repair' the synthetic source sentences. \shortciteA{kumari-etal-2021-domain}   likewise show improvements when adapting NMT on domain-filtered back-translated data.

A related approach proposed by \shortciteA{currey-etal-2017-copied} does not back-translate in-domain target sentences at all, but instead copies them to the source side. This can  teach the model to produce rare words present in  target language in-domain monolingual data without requiring an expensive back-translation model.  \shortciteA{burlot-yvon-2018-using} show that training on such copied sentence pairs with language-tags on the source side can give similar results to using true back translation. 

\shortciteA{zhang-toral-2019-effect} and \shortciteA{graham-etal-2020-statistical} have shown that training on high proportions of back-translations gives a false sense of model quality on back-translated test sets, while the apparent improvements may not be seen on natural test sets. Effectively, back translations can constitute their own domain.  Tagging back translated data, as for a domain (section \ref{sec:doctag}) appears to mitigate this effect \shortcite{caswell-etal-2019-tagged}, and is particularly successful in low-resource scenarios \shortcite{marie-etal-2020-tagged}, further emphasising the domain-like nature of back translation. Adapting to back-translated data  thus risks adapting to this `translationese' domain instead of the true domain of interest. \shortciteA{DBLP:conf/acl/JiaoYSL21} propose mitigating this effect by interposing authentic natural parallel data, if available, when tuning on back-translated data.

\subsubsection{Forward Translation}
\label{sec:forwardtranslation}


Forward translation generates synthetic target sentences with an existing source-to-target NMT model. A sub-type, self-learning, trains a model with its \emph{own} synthetic translations. Forward translation is less common than back translation, perhaps because the synthetic data is on the target side and so any forward translation errors may be reinforced and produced at inference. However, self-learning   can mean more efficient domain adaptation than back translation, as it  requires no additional  NMT model. \shortciteA{chinea-rios-etal-2017-adapting} demonstrate self-learning for domain adaptation, producing data in the target domain by forward translation and then tuning on that data. \shortciteA{zhang-zong-2016-exploiting} find  self-learning particularly beneficial for low-resource target languages.  As in previous sections, relevant monolingual data can be selected using domain-specific terminology or n-gram overlap \shortcite{haque-etal-2020-terminology} or  continuous representations \shortcite{chinea-rios-etal-2017-adapting}. 

A variation of forward translation uses one or more much larger or otherwise stronger `teacher' model to generate in-domain forward translations which are then used to train or tune a `student' model. \shortciteA{gordon-duh-2020-distill} demonstrate that even strong generic models that have undergone this process in the generic domain can benefit from training on in-domain forward translations from a teacher.  \shortciteA{currey-etal-2020-distilling} obtain forward translations from multiple separate in-domain teachers to train a multi-domain system. 

Unlike back translation, forward translation can also be applied to precisely the in-domain source text we actually wish to translate, such as a new  test document. In this scenario we can generate synthetic test target sentences. This can be used for direct fine-tuning, or as a seed to retrieve more natural or synthetic parallel sentences for adaptation as in \shortciteA{poncelas2018data}. In this way forward translation permits few-shot translation to previously unseen or unknown domains in a multi-domain scenario.


Adapting to forward translated data does not risk adaptation to a  `domain' of synthetic inputs as for back translation, since the source sentences remain natural. However, the presence of target-side synthetic sentences  still requires caution, since the model may learn to generate translationese for the domain of interest. Any forward translation errors will also be reinforced by adaptation. Indeed, \shortciteA{DBLP:journals/corr/abs-1911-03362} suggest the outcome of adapting to forward translations depends heavily on the quality of the data-generating system, with humans typically preferring the fluency of systems trained on back translation.

\subsection{Artificially Noising and Simplifying Natural Data}

In some cases only a small in-domain dataset is available for adaptation, with neither bilingual nor monolingual large in-domain corpora from which to extract additional relevant sentences. In this scenario we can generate new data from the available in-domain set by changing its source or target sentences in some way. Including these variations in tuning can  reduce the likelihood of overfitting or over-exposure to a small set of one-to-one adaptation examples \shortcite{bishop1995training}, thus addressing one of the fine-tuning pitfalls of section \ref{sec:defaultdomainadaptation}.

A common example is adding artificial noise to source  sentences, for example by deleting, substituting or permuting characters or words. As well as mitigating overfitting during adaptation, tuning on such sentences can improve NMT robustness. Tuning on noised data can improve performance on new sentences containing natural mistakes \shortcite{vaibhav-etal-2019-improving,karpukhin-etal-2019-training} or noise introduced synthetically, for example by a poorly-performing automatic speech recognition (ASR) system \shortcite{sperber2017robust,digangi2019asr}.  The noised dataset may also be considered a domain in itself, as for the work of  \shortciteA{tan-etal-2020-morphin} adapting NMT to handle a wider range of natural linguistic variation. \shortciteA{kim-etal-2021-using} highlight another benefit of performing domain adaptation with noisy data, in their case heavily segmented sentence fragments. They suggest that sufficient noising may serve to redact any confidential information in adaptation data, mitigating privacy concerns while still benefiting in-domain translation performance. \shortciteA{DBLP:journals/corr/abs-2106-11375} also fine-tune on fragmented phrase translations, but note that fragmentation-noising  eventually limits performance on non-noisy longer sentences.

New in-domain examples can  also be constructed  by simplifying some natural source  \shortcite{li2019explicit,hasler2017source} or target \shortcite{agrawal-carpuat-2019-controlling} sentences. Simplifying sources can make translation easier and can be applied to in-domain test sentences. Simplifying targets allows specification of output language complexity, which may itself be a domain feature. However, we find  simplification approaches to adaptation are rare compared to noising approaches. This may be due to the difficulty in obtaining simplification training data and models, which themselves may  be domain-sensitive or domain-mismatched, while noising text is intuitively often easier. Interestingly, \shortciteA{mehta2020simplify} find that back-translation  can function as a simplification system, and  that the resulting simplified sentences improve NMT in low resource settings. This may suggest that   generating domain adaptation data with back-translation  (section \ref{sec:backtranslation}) is beneficial in part due to the simplified  synthetic source sentences. 


\subsection{Synthetic Bilingual Adaptation Data}
\label{sec:purelysynthetic}
A final type of data used for adaptation is purely synthetic. Purely synthetic data may be beneficial when even monolingual in-domain natural data is unavailable, or when it is not practical to back- or forward-translate such data, perhaps due to lack of a sufficiently strong NMT model. Synthetic data may be obtained from an external or induced lexicon, or constructed from a template. 

Lexicons have been used effectively when dealing with rare words, OOV words or ambiguous words with multiple senses in the training data \shortcite{zhao-etal-2018-addressing}. For SMT lexicons can be used to mine in-domain translation model probabilities directly \shortcite{daume-iii-jagarlamudi-2011-domain}. In NMT domain-specific lexicon probabilities may be incorporated into the NMT loss function \shortcite{arthur-etal-2016-incorporating} or bilingual lexicon entries may be used to construct partially synthetic sentence pairs \shortcite{zhang2016bridging}.

Another application of synthetic lexicon data is covering words or phrases for which there is an easily obtainable translation, and  which the model is likely to be required to translate. This type of data is usually domain specific: NMT for use on social media may require translations for common greetings, while a biomedical NMT system may have a fixed set of terminology to translate. \shortciteA{hu-etal-2019-domain-adaptation} adapt to this form of lexicon for a pre-defined domain, and find that it is complementary to tuning on back translated data, with lexicons aiding translation of isolated words and back translations aiding overall in-domain fluency. \shortciteA{kothur-etal-2018-document} adapt to a lexicon containing novel words in a test document, and find that careful hyperparameter tuning is necessary to mitigate overfitting behaviour when adapting to dictionaries for too many epochs. \shortciteA{DBLP:journals/corr/abs-2004-02577} mitigate this form of overfitting by  combining generic domain parallel data with in-domain dictionaries to synthesize in-domain parallel sentences.

\section{Architecture-Centric Adaptation}
\label{litreview-architecture}
Architecture-centric approaches to domain adaptation typically add trainable parameters to the NMT model itself. This may be a single new layer, a domain discriminator, or a new subnetwork.  If domains of interest are known ahead of model training, domain-specific parameters may be learned jointly with NMT model training. However, in a domain adaptation scenario where we wish to avoid retraining, domain-specific parameters  can be added after pre-training but before any fine-tuning on in-domain data.

Many  of the papers surveyed in this section  describe learning domain-specific labels or parameters when training NMT models from scratch. In that sense their focus is on domain adaptation in the sense of adapting the domain of a specific translation, but \emph{not}  as described in section \ref{litreview-domainproblemdescription}, where we wish to achieve good in-domain performance without training from scratch. We survey these papers regardless for two reasons. One is completeness: many approaches described here have become popular as ways to control or improve domain-specific translation. The second is that  the choice to train from scratch with domain-specific features  may often be based only on relative simplicity of implementation. Editing the architecture of a pre-trained model can be more complex, if more efficient, than simply retraining a new model with a different architecture. We highlight work where domain-specific parameters or other modeling changes are made after pre-training, notably adapter layers (section \ref{sec:adapter}) as indicative that  similar `from scratch' approaches described here could also adapt without requiring retraining.


\subsection{Domain Labels and Control}
\label{sec:doctag}
 \begin{figure}[ht]
\begin{center}
 \includegraphics[scale=0.8]{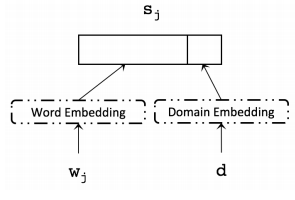}
  \end{center}
 \caption{Illustration of domain labelling from  \shortciteA{kobus2017domain}. A source token embedding is made up of both the source word embedding and a domain embedding.}
 \label{fig:label}
 \end{figure}

Where data is domain-labelled, the labels themselves can be used to signal domain for a multi-domain system. Domain labels may come from a human-curated source or from a model trained purely to infer domain. Tags  can be applied to either the source or target sentence. While tags may be treated as simply another token in the source or target sequence -- `inline' tagging --  they are also frequently generated by a dedicated subnetwork, or involve a separate embedding  (Figure \ref{fig:label}). We therefore discuss tags here in conjunction with architectural considerations. 

Tagging the source allows external control of the target domain. One motivation for source tagging is that some sentences could conceivably be translated into multiple domains, particularly given that domain can encompass genre-related concepts like formality register, as discussed in section \ref{secdomain}. Another motivation is that contemporary NMT systems may not successfully infer domain from the source sentence. Instead, there is some evidence that they struggle with   aspects of language that would indicate domain to a human, such as  lexical ambiguity \shortcite{emelin-etal-2020-detecting} or  formality \shortcite{hovy-etal-2020-sound}. Consequently any available domain information may  help to guide  the NMT system during adaptation.

\shortciteA{kobus2017domain} explore domain-tagging the source sentence for NMT with either a single inline token or as an embedded feature combined with each token embedding  (Figure \ref{fig:label}). \shortciteA{tars2018multi} also  source-tag domains  as either a single tag or as a tag feature, using real known-domain labels as well as those determined by external supervised classifiers and unsupervised clustering schemes. Both \shortciteA{kobus2017domain} and \shortciteA{tars2018multi} find slightly better performance for domain features than  for discrete tags. Both also find that tags improve in-domain translation over simple fine-tuning on the target domain, supporting the view that untagged NMT may not successfully  infer domain.

\shortciteA{chu2017empirical}  introduce in-domain and generic domain source sentence tags during fine-tuning. They find this improves performance on the target domain, but show improved performance on the generic domain without tags, and suggest that tags discourage the model from taking advantage of any overlap between the generic and target domains.  \shortciteA{stergiadis-etal-2021-multi}  account for this problem of overlapping  domains by using multiple domain tags for each source sentence.  \shortciteA{mino-etal-2020-content} likewise use  multiple in-line tags indicating domain and whether data is noised, and show benefits over using single tags.    \shortciteA{wang-bridging-2021} fine-tune on mixed domains, but improve over simple domain tagging by tuning with counterfactual tagged examples, where some generic domain source sentences are also labelled with a best-fit in-domain tag.

While most of the above work focuses on tagging the source sentence, \shortciteA{britz-etal-2017-effective} instead prepend inline domain tags to the target sentence. They suggest that although NMT models may not learn to infer domain without intervention, target tagging can provide that intervention and teach the model to classify the source sentence's domain. Importantly this does not require the domain to be explicitly determined for new, user-generated source sentences. However, it also complicates the process of controlling the desired domain.

Although many of the papers proposing  domain control  apply labels to all sentences when training from scratch, this is not strictly necessary. Indeed  \shortciteA{pham-etal-2021-revisiting}, comparing several previously proposed tagging approaches, note that introducing new domains by specifying new labels at fine-tuning time  is straightforward.

\subsection{Domain-Specific Subnetworks}
\label{subnetwork}
Another architecture-based approach introduces new subnetworks for domains of interest. This could be a domain-specific element of the original NMT architecture (e.g. vocabulary embedding, encoder, decoder), or a domain embedding determined by a  subnetwork not typically found in the Transformer. 

Vocabulary embeddings can be made wholly or partially domain-specific.  This is related to the domain-embedding feature discussed in section \ref{sec:doctag}, but different in that the whole  embedding relates to the vocabulary item, not to a domain tag. For example, \shortciteA{zeng-etal-2018-multi} determine both domain-specific and domain-shared source sentence representations.  \shortciteA{pham2019generic}  change existing word embeddings to have domain-specific features which are activated or deactivated for a given input,   improving lexical choice. \shortciteA{DBLP:conf/acl/LinYYLZLHS20}  cache  keywords for individual users with dedicated subnetworks, effectively tracking user `domain'-specific vocabulary embeddings to combine with the  generic  embeddings. \shortciteA{sato-etal-2020-vocabulary} replace generic translation word embeddings with domain-specific vocabulary embeddings learned by domain-specific language models from monolingual data. \shortciteA{dou-etal-2019-unsupervised} also learn domain-specific embeddings using monolingual target data, but via an auxiliary language modelling task learned jointly with model tuning.

While most work mentioned in  section \ref{sec:doctag} assumes known domain labels or determines them using an external classification model, it is also possible to learn a domain classifier jointly with the NMT model, as in \shortciteA{britz-etal-2017-effective} described above. \shortciteA{lachaux-etal-2020-target} represent the target domain as a latent variable input to a shared decoder. In these cases target domain labels are assumed to be available for training, but at inference time domain labels can be input to the model to control the output domain. Even if the desired domain label is unknown at inference, they demonstrate that multiple sets of outputs can be generated by performing inference with different domain label control. 

The domain can be incorporated into the model architecture in ways other than simply changing the word embeddings. For example, \shortciteA{michel-neubig-2018-extreme} learn domain embeddings for many individual speaker `domains' during fine-tuning. They use these to bias the softmax directly, resulting in small improvements over tuning with domain tags prepended to the target sentence. \shortciteA{DBLP:journals/corr/abs-1906-07978} likewise learn  domain-specific softmax biases for multiple domains, although they do not see improvements over simply appending a domain token to the source sentence.   A domain embedding does not need to be determined directly from the source sentence:  \shortciteA{DBLP:conf/nlpcc/WuZZ19} and \shortciteA{wang-2020-general-to-particular}  both learn  domain-specific representations across multiple domains from the output of the encoder, to feed to a shared decoder. These approaches  encourage domain specificity in parts of the model other than simply the source sentence representation.

An alternative approach is to duplicate encoders or decoders for each domain of interest, although this quickly becomes expensive in terms of added model size and therefore computational and energy cost for training and inference. \shortciteA{gu-etal-2019-improving} use both a shared encoder and decoder for all sentences, as well as domain-specific encoders and decoders for each input domain as determined by a classifier. \shortciteA{jiang-etal-2020-multi-domain}  define multi-head attention networks for each domain. The overall attention in each layer is a weighting function of the different domain attention networks, determined by domain classifiers for each layer.


\subsection{Lightweight Added Parameters}
\label{sec:adapter}

 \begin{figure}[ht]
\begin{center}
 \includegraphics[scale=0.45]{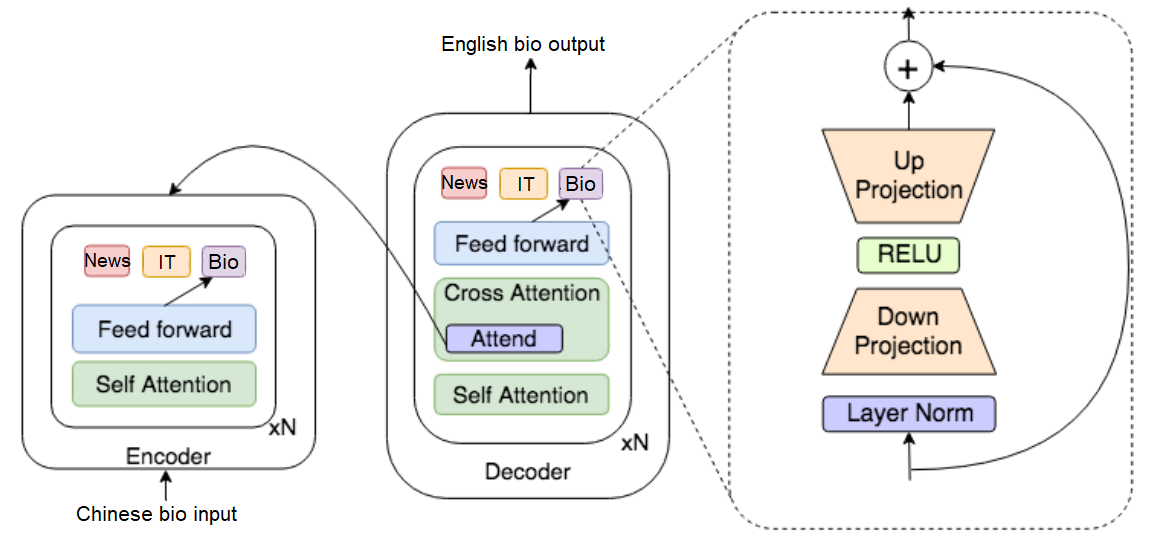}
  \end{center}
 \caption{Illustration of adapter layers used for domain adaptation, based on Figure 1 from \shortciteA{bapna-firat-2019-simple}. A separate set of adapters can be defined for each domain of interest, here News, IT and Bio. The adapter (right) consists of relatively few parameters, and is incorporated into both encoder (left) and decoder (middle) layers.}
 \label{fig:adapter}
 \end{figure}

A lightweight architectural approach to domain adaptation for NMT adds only a limited number of parameters after pre-training. The added parameters are usually adapted on in-domain data while pre-trained parameters are `frozen' -- held at their pre-trained values. For example, \shortciteA{vilar2018learning} introduces a new component to each  hidden unit in the model which can amplify or decrease the contribution of that unit. \shortciteA{bapna-firat-2019-simple} instead add new layers, called `adapters'. In each case the domain-specific multiplicative unit or adapter layer is tuned on the in-domain data while the rest of the model is frozen.

Adapter layers (Figure \ref{fig:adapter}) have achieved  particular popularity as they require very simple and lightweight model modification, and inherently involve no forgetting since the original model parameters are unchanged. They are typically inserted between layers in the encoder and decoder, and may be used alongside a domain discriminator that determines which adapter to use \shortcite{pham-etal-2020-study}.  Notably, \shortciteA{abdul-rauf-etal-2020-limsi} find that adapters can outperform full model fine-tuning when translating a `noisy' domain, to which the full model overfits. However, many of the above papers highlight that using adapters for NMT domain adaptation requires careful choice of adapter size and number of tuning steps.


\subsection{Architecture-Centric Multi-Domain Adaptation}

Architectural approaches as described in this section are capable of good performance over multiple domains. In particular, schemes that leave original parameters  unchanged and only adapt  a small added set of parameters  can avoid any performance degradation from forgetting or overfitting by simply using the original parameters. Adapter-like architectural approaches may therefore have a natural application to continual learning, and can also be a lightweight approach for other multi-domain scenarios.

It is worth noting that using a certain set of parameters for a certain domain implicitly assumes that language domains are discrete, distinct entities. New architecture may be either `activated' if the test set is in-domain, or `deactivated' for better general domain performance as in \shortciteA{vilar2018learning}. A sentence may be assigned to a single domain, and a label added for that domain as in \shortciteA{tars2018multi}. However,  multiple text domains may overlap, and  training domains may be mutually beneficial for translation performance -- using discrete tags may interfere with this, as found by  \shortciteA{chu2017empirical}, and likewise discrete domain-specific subnetworks may not make optimal use of multi-domain data.

\shortciteA{pham-etal-2021-revisiting} compare various architecture-centric approaches to NMT domain specificity  mentioned in this section,  focusing on their performance in a multi-domain setting. They  highlight that systems relying on domain tags  are  hampered by wrong or unknown tags, for example for new domains or unfamiliar examples of existing domains. Among the compared multi-domain  systems, they find performance generally underperforms simple fine-tuning for a single given domain, but that when domains are close or overlapping, multi-domain systems that share parameters across different domains are most effective.

\section{Training Schemes for Adaptation}
\label{litreview-domainforgetting}

Once data is selected or generated for adaptation and a neural architecture is determined and pre-trained, the model can be adapted to the in-domain data. One straightforward approach described in section \ref{sec:defaultdomainadaptation} is fine-tuning the neural model with the same MLE objective function  used in pre-training. However, as mentioned there, simple fine-tuning approaches to domain adaptation can cause catastrophic forgetting of old domains and overfitting to new domains. In this section we  discuss training schemes intended to mitigate these problems: regularized training, curriculum learning, instance weighting and  non-MLE training.

\subsection{Objective Function Regularization}
\label{litreview-domainregularization}

A straightforward way to mitigate forgetting is to minimize changes to the model parameters. Intuitively, if parameters stay close to their pre-trained values they will give similar performance on the pre-training domain. Here we discuss  regularization and knowledge distillation techniques which may be applied during training for this purpose.

\subsubsection{Freezing Parameters}
\label{sec:freeze}
One way to ensure model parameters do not change is to simply not update them, effectively dropping them out of adaptation (section \ref{sec:distribreg}). Research has shown that effective adaptation results can be obtained by varying just a small subset of the original model parameters\footnote{Different to adapters (section \ref{sec:adapter}), which introduce \emph{new} parameters and freeze all original parameters.}.  For example, \shortciteA{thompson2018freezing} simply choose subsets of recurrent NMT model parameters to hold at their pre-trained values when fine-tuning on a new domain.  \shortciteA{gu-feng-2020-investigating} extend this work to the Transformer architecture.  Both note a decrease in forgetting dependent on which subnetwork is frozen, although performance on the new domain is reduced relative to fine-tuning. \shortciteA{wuebker-etal-2018-compact} likewise adapt only  a manually defined subset of model parameters, encouraging sparsity in the adapted parameters with L1 regularization to improve efficiency. \shortciteA{deng-etal-2020-factorized} factorize all model components into shared and domain-specific, freezing the shared factors when tuning the domain-specific components on multiple domains. In these two cases performance on the new domains is shown to be similar to full fine-tuning, even though only a fraction of the parameters are adapted. 

Parameters to be frozen do not necessarily have to be defined manually. \shortciteA{DBLP:conf/aaai/LiangZWQ021} identify low-magnitude parameters as sparse or underused areas of the network that can be easily adapted to new domains. They prune the sparse areas, then tune the original network  on the generic domain so that it recovers, after which the sparse areas can be reintroduced and tuned on the new domain while the rest of the network is frozen.  \shortciteA{gu2021pruning} take a very similar approach, but recover the pruned model using knowledge distillation with in-domain data, removing the need for access to the generic dataset. In both cases the authors demonstrate adaptation to multiple domains simultaneously while significantly reducing forgetting on the generic domain.

As for the  approaches described in section \ref{sec:adapter}, these schemes have the advantage that only a small number of parameters need be saved for each adapted model. This is particularly useful for saving memory if `personalizing' many models for multiple fine-grained domains, such as individual users.

\subsubsection{Regularizing Parameters}
\label{sec:regularize}
 \begin{figure}[ht]
\begin{center}
 \includegraphics[scale=0.65]{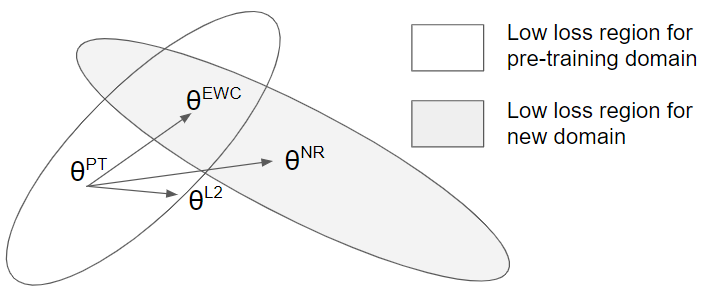}
  \end{center}

 \caption{Illustration of parameter regularization during fine-tuning from a pre-training (PT) domain to a new domain, based on Figure 1 from \shortciteA{kirkpatrick2017overcoming}. If the pre-trained parameters $\theta^{PT}$ are adapted with no regularization (NR), good performance on the new domain corresponds to catastrophic forgetting on PT. Applying the same regularization to all parameters (L2) encourages minimal overall change from $\theta^{PT}$. EWC regularization \shortcite{kirkpatrick2017overcoming} aims to vary the parameters that are unimportant for PT.}
 \label{fig:ewc-l2}
 \end{figure}

If adapting all parameters, regularizing the amount of adaptation can mitigate forgetting. For example, \shortciteA{barone2017regularization} allow all NMT model parameters to vary under L2 regularization relative to their pre-trained values $\theta^{PT}$.  \shortciteA{kirkpatrick2017overcoming} introduce the related approach of Elastic Weight Consolidation (EWC) for computer vision domain adaptation, which effectively scales the L2 regularization applied to each parameter $\theta_j$ by a value $F_j$.   If $\Lambda$ is a scalar weight,  a general-form L2 regularized loss function is:    
\begin{equation}
\hat{\theta} = \argmin_{\theta} [L_{CE}(\mathbf{x}, \mathbf{y}; \theta) + \Lambda \sum_j  F_j  (\theta_j - \theta^{PT}_j)^2 ]
\label{eq:l2}
\end{equation}
We illustrate these approaches in Figure \ref{fig:ewc-l2}.
Relative domain importance can be controlled or tuned with $\Lambda$, which is larger if the old domain is more important and smaller if the new domain is more important. L2 regularization occurs where $F_j = 1$ for all $j$. For EWC \shortciteA{kirkpatrick2017overcoming}  define $F_j$ as the Fisher information for the pre-training domain estimated over a sample of data from that domain, $(\mathbf{x}^{PT}, \mathbf{y}^{PT})$.

\begin{equation}
F_j = \mathbb{E}\big[ \nabla^2  L_{CE}(\mathbf{x}^{PT}, \mathbf{y}^{PT};\theta^{PT}_j)\big]
\label{eq:ewc}
\end{equation}


EWC has been broadly applied to NMT domain adaptation in recent years. \shortciteA{thompson2019ewc} demonstrate that it mitigates  catastrophic forgetting adapting from a generic to a specific domain. \shortciteA{saunders-etal-2019-domain} demonstrate that this extends to sequential adaptation across multiple domains, and \shortciteA{stahlberg-etal-2019-cued} show that EWC can improve translation on the generic domain when adapting between related domains. 

\shortciteA{cao-etal-2021-continual} observe that catastrophic forgetting corresponds to the model reducing the probability of  words from previous domains by shrinking their corresponding parameters. They address this by normalizing target word parameters, an approach previously proposed by \shortciteA{nguyen-chiang-2018-improving} to improve rare word translation by mitigating the bias towards common words. While not quite the same as regularizing parameters via the objective function, the immediate goal of avoiding excessive changes in parameter magnitude during tuning is the same, as is the effect of reducing forgetting.

Parameter regularisation methods involve a higher computational load than parameter freezing,  since all parameters in the adapted model are likely to be different from the baseline, and the baseline parameter values are often needed for regularisation. However, regularisation may also function more predictably across domains, language pairs and models than selecting parameters to freeze by subnetwork.
\subsubsection{Knowledge Distillation}
\label{sec:KD}
Knowledge distillation and similar `teacher-student' model compression schemes effectively use one teacher model to regularize training or tuning of a separate student model \shortcite{bucilua2006model,Hinton2015DistillingTK}.  Typically the teacher model is a large, pre-trained model, and the student is required to emulate its behaviour with far fewer parameters. The student model is tuned so its \emph{output distribution} remains similar to the teacher's output distribution over the pre-trained data, often via an addition to the loss function \shortcite{kim-rush-2016-sequence}. 

In a domain adaptation context, knowledge distillation encourages similar performance on the pre-training domain with a regularization function between general and in-domain model output distributions, with the teacher being another NMT model. \shortciteA{dakwale2017fine} focus on weighting the teacher distribution weighting to reduce catastrophic forgetting, while  \shortciteA{khayrallah2018regularized} aim for better in-domain performance on a single domain and \shortciteA{DBLP:conf/ecai/MghabbarR20} use knowledge distillation to adapt to multiple domains simultaneously. \shortciteA{cao-etal-2021-continual} extend  knowledge distillation for domain adaptation to a continual learning scenario where the model is adapted sequentially multiple times. They propose dynamic knowledge distillation, in which older versions of the model carry less weight compared to newer versions.

We see knowledge distillation as similar in spirit to parameter regularization. However, it has a higher computational cost, since two models must actually operate on the data, instead of some parameter values simply being stored in memory. This can be effective when the aim is to compress the teacher model, since in this case the in-domain student model is likely to be much smaller than the other. For models remaining the same size, parameter regularization may be more practical. 


\subsection{Curriculum Learning} 
\label{litreview-curriculum}
 \begin{figure}[ht]
\begin{center}
 \includegraphics[scale=0.5]{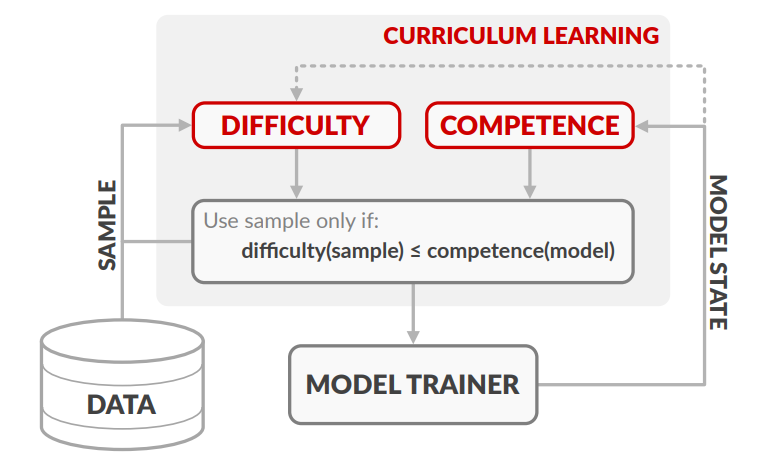}
  \end{center}

 \caption{Illustration of competence-based curriculum learning from  \shortciteA{platanios-etal-2019-competence}. A training example is used only if its difficulty is within the model's competence. Perceived example difficulty may also be dependent on the model competence.}
 \label{fig:curriculum}
 \end{figure}

\shortciteA{bengio2009curriculum} recognize that humans learn  best when concepts are presented in a meaningful order, or a `curriculum'. They hypothesize that neural model training can benefit from the same strategy of curriculum  learning in terms of learning speed or  quality, and demonstrate that this is the case for language modelling with a curriculum of increasing vocabulary.  In broad terms, a curriculum ranks the training examples. The ranking guides the order in which examples are presented to the model during training or fine-tuning. A typical curriculum orders training examples by difficulty, with the easiest examples shown first and the more complex examples introduced later \shortcite{zhang-etal-2017-boosting,weinshall2018curriculum}. Difficulty can be determined in terms of objective data features like sentence length, linguistic complexity or word rarity \shortcite{kocmi-bojar-2017-curriculum,platanios-etal-2019-competence}. Although an easiest-first ranking is common, other rankings are possible. In fact, \shortciteA{zhang2018empirical} find that both easy-first and hardest-first schedules can give similar speed improvements.

Figure \ref{fig:curriculum} illustrates a general form of curriculum learning in which both example difficulty and model `competence' determine whether a given training example is used at a certain point in training. Difficulty and competence may both be estimated directly from the model  \shortcite{platanios-etal-2019-competence}. For example, a training example may be considered  difficult for an NMT model at a given point in training if the example's embedding norm is large \shortcite{liu-etal-2020-norm}, if the training loss of the example is changing significantly between training iterations \shortcite{wang-etal-2018-dynamic}, or if the model simply does not translate it well \shortcite{dou2020dynamic}. We note that determining model competence by translating all training examples during training or tracking competence across  epochs may increase the computational cost to an extent that outweighs any benefits of better data ordering.

More relevant to this survey, a curriculum  can be constructed from least-domain-relevant examples to most-domain-relevant. In fact, simple fine-tuning (section \ref{sec:defaultdomainadaptation}) is effectively curriculum learning where the final part of the curriculum contains only in-domain data.  
However, curriculum-based approaches to domain adaptation generally involve a gradual transition to in-domain data. For example, \shortciteA{wang-etal-2018-denoising} use an incremental denoising curriculum to fine-tune a pre-trained NMT model on increasingly clean data from its existing training corpus. Similar `cleaning' fine-tuning curricula data can be learned via reinforcement learning methods \shortcite{kumar-etal-2019-reinforcement,zhao2020reinforced}.


When adapting to a distinct domain a curriculum can be determined in terms of similarity with either target domain data or with generic domain data. Moving smoothly between the generic and target domains may mitigate forgetting and overfitting. One way to move smoothly from the generic to the target domain is to extract target-domain-relevant data from  generic data already seen by the model. For example, \shortciteA{zhang-etal-2019-curriculum} identify `pseudo in-domain' data from general corpora and use similarity scores to rearrange training sample order. This effectively treats distance from in-domain data as a curriculum difficulty score.  \shortciteA{van-der-wees-etal-2017-dynamic} likewise perform domain adaptation while gradually emphasising more in-domain data from the generic dataset each training epoch. 

The idea of a data sampling curriculum can be extended to multi-domain models. \shortciteA{sajjad2017neural} carry out `model stacking', training on incrementally closer domains to the target domain, spending several epochs on each domain with the aim of maintaining good performance across multiple domains.  \shortciteA{wang-etal-2020-learning-multi} use a difficulty-based domain adaptation curriculum across multiple domains simultaneously, using multi-dimensional domain features as difficulty scores.  \shortciteA{wu-etal-2021-uncertainty} use  model uncertainty on a small, trusted multi-domain dataset to determine a curriculum across  corpora for a multi-domain model.

A different perspective on on curricula for domain adaptation `reminds' the model of the generic domain during adaptation, if generic data is available at adaptation time. Incorporating previously-seen data into later points of the curriculum can  reduce forgetting and overfitting, since the model is tuned on a more diverse dataset. For example, \shortciteA{chu2017empirical} propose mixed fine-tuning, which incorporates some amount of domain-tagged generic data into the adaptation data for the target data. They demonstrate strongly reduced catastrophic forgetting compared to simple fine-tuning, and complementary behaviour to domain tagging (section \ref{sec:doctag}). 
\shortciteA{Song2020DomainAO} extend mixed fine-tuning to a multi-stage multi-domain curriculum, sequentially learning new domains while discarding less relevant prior-domain data from the data mix, and show that this improves over single-stage mixed fine-tuning. \shortciteA{haslerimproving} demonstrate that mixed fine-tuning without domain tags is complementary to directly regularizing parameters using EWC (section \ref{sec:regularize}). `Reminder' techniques can  be effective even if domains are not treated as distinct entities:  \shortciteA{aljundi2019gradient} demonstrate that maintaining a representative `replay buffer' of past training examples  avoids forgetting even without hard domain boundaries  available.

\subsection{Instance Weighting}
\label{litreview-instance}

Instance weighting adjusts the loss function to weight training examples  according to their target domain relevance  \shortcite{foster-etal-2010-discriminative}. For NMT, an instance weight $W_{x,y}$ for each source-target training example can easily be integrated into a cross-entropy loss function:

\begin{equation}
L(\mathbf{x}, \mathbf{y}; \theta) = \sum_{(\mathbf{x, y})}-W_{\mathbf{x, y}} \log P(\mathbf{y}| \mathbf{x}; \theta)
\end{equation}

A higher weight indicates that a sentence pair is  important for training towards the target domain, while a low (or zero) weight will lead to sentences effectively being ignored during tuning. While this involves adjusting the loss function rather than data reordering, the effect is to simulate over- or under-sampling particular examples during adaptation.

The weight may be determined in various ways. It may be the same for all sentences marked as from a given domain, or defined for each sentence using a domain  measure like n-gram similarity \shortcite{joty-etal-2015-avoid}  or  cross-entropy difference \shortcite{wang-etal-2017-instance}. If changes can be made to the model architecture, the instance weight may be determined by a domain classifier \shortcite{chen-etal-2017-cost}, or an architecture-dependent approach like sentence embedding similarity \shortcite{zhang-xiong-2018-sentence}. The same effect  is achieved by \shortciteA{farajian-etal-2018-eval} by adjusting the learning rate and number of adaptation epochs dedicated to  tiny adaptation sets based on their domain similarity. In each case the outcome is to place more or less emphasis on particular adaptation examples during training.

We view instance weighting as fundamentally the same idea as curriculum learning (section \ref{litreview-curriculum}). Both schemes bias the model to place more importance on certain training examples, allowing some control over how the model fits, forgets, or overfits certain sentences. Some forms of curriculum learning are implemented in a similar way to instance weighting, with a higher weight applied to examples that fall into the current section of the curriculum, or a zero weight applied to examples that should not yet be shown to the model \shortcite{bengio2009curriculum,dou2020dynamic}. One difference is that instance weights for domain adaptation do not usually change as training progresses or model competence changes, but bias the model towards in-domain data in a constant manner.


\subsection{Non-MLE Training}
\label{litreview-mrt}

As discussed in section \ref{litreview-nmttraining}, MLE training is particularly susceptible to exposure bias, since it tunes for high likelihood only on the sentences available in the training or adaptation corpus. MLE also experiences loss-evaluation metric mismatch, since it optimizes the log likelihood of training data while machine translation is usually evaluated with translation-specific metrics. Tuning an NMT system with a loss function other than the simple MLE on pairs of training sentences may therefore improve domain adaptation. Here we describe two other training objectives that have been demonstrated as useful precursors or follow-ups to domain-adaptive MLE fine-tuning: minimum risk training and meta-learning.

\subsubsection{Minimum Risk Training}
Discriminative training for MT was introduced for phrase-based machine translation, minimizing the expected cost of model hypotheses with respect to an evaluation metric like document-level BLEU \shortcite{papineni-etal-2002-bleu}. \shortciteA{shen2016minimum} extend these ideas to Minimum Risk Training (MRT) for NMT, using expected minimum risk at the sequence level with a sentence-level BLEU (sBLEU) cost  for end-to-end NMT training.  Given $N$ sampled target sequences $\mathbf{y}_n^{(s)}$ and the corresponding reference sequences  $\mathbf{y}^{(s)\ast}$ for the $S$ sentence pairs in each minibatch, the MRT objective is: 
\begin{equation}
\hat{\theta} = \argmin_{\theta}\sum_{s=1}^S \sum_{n=1}^N   \Delta(\mathbf{y}_n^{(s)}, \mathbf{y}^{(s)\ast}) \frac{P(\mathbf{y}_n^{(s)}|\mathbf{x}^{(s)};\theta)^\alpha}{\sum_{n'=1}^N P(\mathbf{y}_{n'}^{(s)}|\mathbf{x}^{(s)};\theta)^\alpha}
\end{equation}

Hyperparameter $\alpha$ controls the smoothness of the sample probability distribution. Function $\Delta(.)$ measures hypothesis cost $\in [0, 1]$, typically $1 - \text{sBLEU} (\mathbf{y}_n^{(s)}, \mathbf{y}^{(s)\ast})$. Model hypotheses  $\mathbf{y}_n^{(s)}$ can be generated by beam search or by autoregressive sampling. MRT has been applied to NMT in various forms since its introduction. \shortciteA{edunov2018classical} explore variations on MRT, using samples produced by beam search and an sBLEU-based cost. \shortciteA{wieting-etal-2019-beyond} use MRT  with sBLEU and  sentence similarity metrics. \shortciteA{saunders2020contextobjective} use MRT for translation with BLEU calculated over minibatch-level `documents'. 

MRT is of particular relevance to domain adaptation for two  reasons. Firstly, in the NMT literature we find that MRT is exclusively applied to fine-tune a model that has already converged under a maximum likelihood objective. MRT therefore fits naturally into a discussion of improvements to pre-trained NMT models via parameter adaptation. 

Secondly, there is some indication that MRT may be effective at reducing the effects of exposure bias. Exposure bias  can be a particular difficulty where there is a risk of overfitting a small dataset, which is often the case for domain adaptation, especially if there is a domain mismatch between adaptation and test data \shortcite{muller-etal-2020-domain}. However, MRT optimizes with respect to model samples rather than tuning set target sentences -- the latter  may be less diverse and thus easier to overfit. Moreover, a high-quality NMT baseline may produce synthetic samples that are better aligned than available natural data. The ability of MRT to mitigate overfitting  is highlighted by  \shortciteA{neubig-2016-lexicons}, who notes that MRT tends to produce sentences of the correct length without needing length penalty decoding.  When applied to adaptation, MRT has been shown to mitigate exposure bias when the test data domain is very different from the training domain \shortcite{wang-sennrich-2020-exposure}, and when adaptation data is in-domain but noisy, poorly aligned or unreliable \shortcite{saunders-etal-2020-exposure}.
\subsubsection{Meta-Learning}
\label{litreview-metalearn}

Meta-learning was proposed by \shortciteA{DBLP:conf/icml/FinnAL17} as a means of tuning a neural model such that it can easily learn new tasks in few additional steps: `learning to learn'. The aim is to find model parameters that can be adapted to a range of domains easily, in few training steps and with few training examples. This is achieved by introducing a meta-learning phase after pre-training and before fine-tuning on any particular target domain.

During the meta-learning phase the model $f_\theta$  sees a number of few-shot adaptation tasks $T$, where each task consists of training and test examples. In a meta-training step, model parameters $\theta$ are first temporarily updated using the training examples, then the updated model's loss $L_T$ on the task test examples is used as the real meta-training objective. Effectively, test losses for the meta-training tasks are used as the model's training loss during meta-training.
 For a given learning rate $\alpha$:
\begin{equation}
 \argmin_{\theta}\sum_{T}L_T(f_{\hat{\theta}}) = \sum_{T}L_T(f_{\theta - \alpha \nabla_{\theta}L_T(f_{\theta})}) 
\label{eq:metalearn}
\end{equation}

\shortciteA{sharaf-etal-2020-meta} apply meta-learning to NMT domain adaptation. They specifically meta-tune the parameters of a single adapter layer (section \ref{sec:adapter}) and simulate small domains for meta-training by sampling sets of just hundreds of parallel sentences. They find that this outperforms simply fine-tuning on the same samples when the number of simulated domains is high, but not when fewer, larger meta-training domains are sampled, potentially due to overfitting the meta-training sets. They also note that meta-training all Transformer parameters instead of an adapter  leads to catastrophic forgetting of the original domain. \shortciteA{DBLP:conf/icann/SongMQL21} also meta-learn adapter layers, but adjust meta-learning learning rate $\alpha$ dynamically to make the process sensitive to domain differences. They focus on domain differences in terms of model confidence in modelling a particular domain, and how representative each sentence pair is of a particular domain. They find dynamic learning rate adjustment particularly improves meta-learning performance when target domains of interest are very different sizes. \shortciteA{DBLP:conf/acl/ParkTKYKPC20} apply meta-learning to unsupervised domain adaptation with only monolingual data. They do so by incorporating back-translation and language modelling objectives into the meta-learning phase, and demonstrate strong improvements over both simple fine-tuning and mixed fine-tuning.

\shortciteA{DBLP:conf/aaai/LiWY20}  meta-learn the encoder and vocabulary embedding spaces, and find this outperforms fine-tuning except for domains with many  infrequent words. \shortciteA{zhan-metacurriculum-2021} effectively apply a curriculum section \ref{litreview-curriculum}) to the meta-learning process. They initiate the meta-learning phase by sampling similar meta-training domains, and gradually introduce more divergent domains. They find small improvements for the domains  seen during meta-learning. The results of these two papers together suggest that meta-learning has the greatest potential to  out-perform fine-tuning for variations on domains where  test data is well-represented by the tuning data.


\section{Inference Schemes for Adaptation}
\label{sec:review-inference}

One way to side-step the problem of translating multiple domains in a single model is to simply assign a separate NMT model to each domain and combine them at inference time. Such models can be obtained using the techniques discussed in the previous  sections, for example by fine-tuning a single pre-trained model on data from each domain of interest. While this approach is simple, if not memory-efficient, it begs the question of how best to perform translation on an unseen source sentence from an unknown domain. Possible  approaches are multi-domain ensembling, and reranking or rescoring an existing set of  translation hypotheses. 

We also describe ways to encourage a model to produce domain-specific terminology or phrasing via a pre- or post-processing step at inference time. This approach bypasses even domain-specific tuning of an NMT model.


%
%
\subsection{Multi-Domain Ensembling}
\label{sec:review-multidomainensemble}
At inference time an NMT ensemble can use predictions from multiple models to produce a translation in a single pass, as described in section \ref{litreview-nmtinference}. Here we discuss two forms of domain-specific ensembling. Domain adaptive ensembling seeks to find a good set of interpolation weights for a traditional ensemble of NMT models when translating a sentence of unknown domain. Retrieval-based ensembling is a recently-proposed technique that interpolates NMT predictions with a distribution over  tokens that are likely in context  from a static in-domain datastore.

\subsubsection{Domain Adaptive Ensembling}
Certain models in an ensemble may be more useful than others for certain inputs. For example, if ensembling a software-domain model with a science-domain model, we might expect the science model to be more useful for translating medical abstracts. This idea of varying utility across an ensemble is particularly relevant when the domain of a test sentence is unknown and therefore the best ensemble weighting  must be determined at inference time. We make an important distinction between the domain of the training data and model – typically known, often with an available development set – and the test data, for which we do not always  know the domain or have access to a development set.

With a source sentence of unknown domain, \shortciteA{freitag2016fast} demonstrate that reasonable performance can be achieved via a uniform ensemble of general models and in-domain translation models. However, this seems non-optimal: intuition would suggest that the in-domain model should be prioritised when inputs are closer to in-domain data, and the general model if inputs are more similar to generic training data. Uniform ensembling is straightforward, but does not emphasise performance on any particular domain.

\shortciteA{sajjad2017neural} also use multi-domain ensembles, but weight the contribution of each ensemble model as in Eq. \ref{eq:weight-ensemble}. They determine static ensemble weights tuned on development sets.   This static weighting approach can be targeted to a particular domain, but weight tuning can be very slow, and  requires either that the domain of the test set is known or that a development set representative of the test set is available.
 \begin{figure}[ht]
\begin{center}
 \includegraphics[scale=0.65]{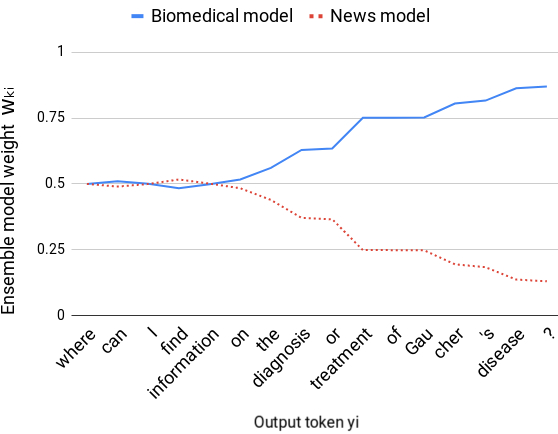}
  \end{center}
 \caption{Bayesian Interpolation for multi-domain ensembling from  \shortciteA{saunders-etal-2019-domain}. Interpolation weight $W_k$ for each of the $K=2$ domain models  is adjusted at each inference step $i$ depending on the likelihood of the hypothesis so far under each model.}
 \label{fig:biensemble}
 \end{figure}

Domain adaptive approaches to inference can instead determine the ensemble weights conditioned only on the source sentence or the partial translation hypothesis, without the need to tune weights manually. For SMT, \shortciteA{huck2015mixed} use a language model to classify the domain of a test sentence when determining which set of parameters to use when generating the translation hypothesis. \shortciteA{DBLP:journals/ijcse/LiuZTWS20} use text similarity between the test sentence and in-domain corpus to determine an ensemble interpolation weight between generic and in-domain NMT models.   An alternative approach, Bayesian Interpolation, is introduced by \shortciteA{allauzen2011bayesian} for language model ensembling. Bayesian Interpolation allows adaptive weighting without necessarily specifying that the `task', $t$ -- the domain of the test sentence -- corresponds to exactly one of the model domains $k$. Neither is it assumed that a uniform weighting of all $K$ domains is optimal for all tasks. Instead a set of tuned or approximated ensemble weights $\lambda_{k,t}$ defines a task-conditional ensemble:
\begin{equation}p( \mathbf{y} | t) =  \sum_{k=1}^K \lambda_{k,t} \; p_k(\mathbf{y})\end{equation}
This can be used as a fixed weight ensemble if task $t$ is known, much like the approaches described in \shortciteA{huck2015mixed} and \shortciteA{DBLP:journals/ijcse/LiuZTWS20}. However if $t$ is not known, the ensemble can be written as follows at inference step $i$, where $h_i$ is history $\mathbf{y}_{1:i-1}$:
\begin{align}
         p(y_i|h_i) &= \sum_{t=1}^T p(t, y_i|h_i) \notag = \sum_{k=1}^K  p_k(y_i|h_i)  \sum_{t=1}^T p(t|h_i)  \lambda_{k,t}  \notag \\
        &= \sum_{k=1}^K  W_{k,i} \, p_k(y_i|h_i)  
\label{eq:adaptdecode}
 \end{align}
 
That is, a weighted ensemble with state-dependent mixture weights computable from task priors and the updated language model task posterior:
\begin{equation}p (t|h_i) = \frac{p(h_i|t) p(t)} {\sum_{t'=1}^T p(h_i|t') p(t')}\label{eq:task-posterior}\end{equation}

\shortciteA{saunders-etal-2019-domain} extend this formalism to include conditioning on a source sentence. This permits domain adaptive NMT with multi-domain ensembles when the test sentence domain is unknown and may vary within a given input, as illustrated in Figure \ref{fig:biensemble}. If a new, unseen domain is present, an adaptive weighting will favour the models under which the new domain is most likely.   \citeA{saunders-etal-2019-ucam} emphasise that this approach is most beneficial when ensembling models which specialize in non-overlapping domains, and that an ensemble of similar models, each handling multiple domains, may indeed perform well with a uniform weighting.



\subsubsection{Retrieval-Based Ensembling}

An ensemble does not need to consist purely of NMT models, or even involve an additional neural network. \shortciteA{DBLP:conf/iclr/KhandelwalFJZL21} propose $k$-nearest-neighbour ($k$NN) machine translation using a domain-specific data-store. The data-store maps NMT decoder states to in-domain target language tokens. At each inference step the NMT model's predictions are interpolated with a distribution over the $k$ retrieved nearest neighbour tokens given the current model state. This is effectively an ensemble with a distribution from a static context-to-token map. 

\shortciteA{zheng-etal-2021-adaptive} demonstrate that this technique can benefit from determining $k$ itself  adaptively for a given context: retrieving many tokens from rare contexts is likely to result in noisy retrieved neighbours. \shortciteA{DBLP:conf/emnlp/ZhengZHCXLC21} further show that the $k$NN data-store itself can be adapted for a given domain in the absence of parallel data using copying and backtranslation (section \ref{sec:syntheticdata}) to create pseudo-parallel data, and adapters (section \ref{sec:adapter}) to learn states from the pseudo-parallel data. All these variants on $k$NN have been shown to improve in-domain translation  significantly without requiring an additional NMT model. 


\subsection{Constrained Inference and Rescoring}
\label{sec:domainrescore}

\begin{figure}[ht]
\begin{center}
 \includegraphics[scale=0.7]{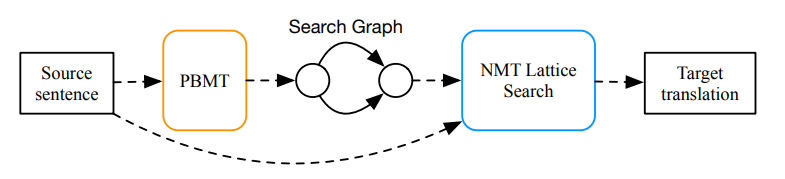}
  \end{center}
 \caption{Illustration of constrained rescoring  \shortciteA{khayrallah-etal-2017-neural}. A domain-specific phrase-based machine translation (PBMT) model is used to create a lattice, which then constrains the NMT output.}
 \label{fig:rescore}
 \end{figure}
Ensembling uses multiple models to produce a translation simultaneously. An alternative is a multi-pass approach to inference, in which one model produces an initial translation which is adjusted or corrected using another model. While this involves multiple models performing their own separate inference passes, this can be more efficient than ensembling. Multi-pass approaches do not involve holding multiple models in memory at once, and the second translation pass is commonly held close to the initial translation in some way, reducing the number of translations to be scored.

The initial model may produce multiple translations -- for example, the highest scoring $N$ translations following beam search (section \ref{litreview-beamsearch}). In this case a second inference pass can rescore this N-best list using a different model or loss function.  For example, Minimum Bayes Risk (MBR) decoding rescores N-best lists or lattices to improve single system performance for SMT \shortcite{kumar-byrne-2004-minimum,tromble-etal-2008-lattice,de-gispert-etal-2010-hierarchical},  or for NMT if a sufficiently diverse lattice can be defined, for example, from SMT n-gram posteriors \shortcite{stahlberg-etal-2017-neural}.  In the context of domain adaptation, \shortciteA{dou-etal-2019-domain} propose rescoring generic NMT model translations with domain-specific language models.    \shortciteA{zhang-etal-2018-guiding} retrieve 
in-domain source sentences that are similar to test inputs, and up-score initial NMT hypotheses which contain fragments of their translations.


A related idea which has been applied to domain-specific translation is constrained inference (Figure \ref{fig:rescore}), where the NMT model is only permitted to produce certain translations. The constraints may be generated by another translation model. For example, \shortciteA{stahlberg-etal-2016-syntactically} generate translation hypotheses with an SMT model, which are then represented as a lattice that constrains NMT decoding.  \shortciteA{khayrallah-etal-2017-neural} constrain NMT output to a domain-specific lattice to improve adequacy in domain adaptation scenarios.  The constraints may allow only certain domain-relevant words to change. For example, \shortciteA{saunders2020genderbias} use lattice constraints when rescoring translations with an adapted model to correct mistakes in a gender `domain'.  They demonstrate that this mitigates forgetting under a domain-adapted model, as the adapted model may only change the original translation  in pre-determined permitted ways. 

The initial beam search may itself be constrained to domain-specific language.  \shortciteA{hokamp-liu-2017-lexically} and \shortciteA{hasler-etal-2018-neural} both adjust the beam search algorithm to force  domain-specific terminology into the output of a generic model, although both approaches are far slower than straightforward beam search.   An extension of the  \shortciteA{hokamp-liu-2017-lexically} approach  by \shortciteA{xie-way-2020-constraining} finds that using alignment information can significantly improve the efficiency of  constrained beam search, with a similar BLEU score.


We note that the initial hypotheses used for rescoring or with constraints do not have to be generated by a domain-specific model, or even a model which the user can tune. For example,  a user might generate initial translations with a generic commercial tool,  then  rescore or retranslate the translations with constraints relating to the desired output, as in \shortciteA{saunders2020genderbias}. It may be  more practical for a user to edit or repair small aspects of generic translations from commercial systems in domain-specific ways, manually or with their own monolingual tool, than to access a custom NMT system.

\subsection{Domain Terminology Via Pre- and Post-Processing}
\label{sec:prepostprocess}
A similar idea to domain terminology `correction' during inference treats the presence of domain-specific terminology as a pre- and post-processing problem. We discuss two such approaches here. The first incorporates domain-specific terminology into the source sentence in pre- and post-processing steps so as to ensure they are correctly translated. The second  preprocesses the input with similar source sentences or translations as a cue for the model.

\subsubsection{Terminology Tagging}
\label{ref:termtag}
There are many possible ways to integrate terminology into the system at inference time. \shortciteA{dinu-etal-2019-training} add inline tags to terminology in the source sentence as a pre-processing step, and then replace the translated tags with corresponding target language terms. \shortciteA{song-etal-2019-code} simply replace certain source phrases with pre-specified domain-specific target translations as a preprocessing step, encouraging copy behaviour.	 \shortciteA{michon-etal-2020-integrating} compare several variations on  inline terminology tags, and find that the approach of \shortciteA{dinu-etal-2019-training} may give the best scores where terminology is already reasonably well-handled, but that it is less effective at introducing poorly-handled terminology. By contrast with the above approaches \shortciteA{dougal-lonsdale-2020-improving} inject terminology after inference  as  a post-processing step using source-target alignments. This has the  advantage of not requiring a model to handle tags, and could in principle be used to introduce terminology to a generic or commercial system output. It is, however, reliant on an effective alignment model.

 \shortciteA{DBLP:conf/ijcai/ChenCWL20} take a similar approach to \shortciteA{dinu-etal-2019-training} and \shortciteA{song-etal-2019-code}, but do not require alignments and only require bilingual dictionaries during inference - they specify the reference terminology in a fixed source position, encouraging the model to learn correct alignments. 	\shortciteA{niehues-2021-continuous} also specify the reference terminology in a fixed source position, but additionally use only the lemma, encouraging the model to learn correct inflections for provided terminology. 
 \shortciteA{DBLP:conf/acl/LeeYC20} add an objective to predict the span of masked source terms during NMT training, allowing them to insert multi-word domain-specific terms during inference. They find similar unigram performance to \shortciteA{DBLP:conf/ijcai/ChenCWL20}, but better reproduction of longer terms. 
 
 We note that many of these approaches do not actually require the baseline model to see or produce the domain-specific terminology. This means terminology injection can take place even for new, unseen domains whose terminology which would not otherwise be generated by the model \shortcite{michon-etal-2020-integrating}.


\subsubsection{In-Domain Priming}
\label{ref:prime}

An extension to priming the model with individual terminology incorporates entire related sentences. For a given sentence, similar source sentences and their translations can be extracted from a parallel dataset using fuzzy matching as described in section \ref{sec:selectdata}. The similar sentence or its translation can be used as a domain-specific input prompt.

\shortciteA{bulte-tezcan-2019-neural} propose priming in this way with Neural Fuzzy Repair (NFR) based on token edit distance: they  incorporate known translations of similar sentences into the model input to act as context cues at inference. \shortciteA{xu-etal-2020-boosting} perform sentence priming experiments with a wider range of similarity metrics, including  continuous representations. They  distinguish between related and unrelated target words, bringing their work closer to terminology tagging approaches of section \ref{ref:termtag}.  \shortciteA{tezcan2021towards} develop NFR towards use of subword representations for identifying matches, and also mark related and unrelated target terms. By contrast to these approaches \shortciteA{pham-etal-2020-priming} `prime' using both source and target sides of related sentence pairs, forcing the model to produce an in-domain target-language `cue' before translating the sentence of interest. Interestingly,  \shortciteA{moryossef-etal-2019-filling} successfully prime  commercial MT systems with short phrases relating to gender or number, suggesting  custom content injection may even be possible in some black-box scenarios. 

Pre- and post-processing approaches may work best if the NMT model is trained from scratch such that it `sees' examples of generic terminology tags or priming cues during training. However, importantly, this training is not domain-specific, and the model does not need to be \emph{retrained} for performance on a particular domain.  At inference time, these approaches effectively adapt a translation's domain-specificity without the need to retrain or further change the model itself.

\section{Case Studies for (Multi-)Domain Adaptation}
\label{litreview-casestudy}
We conclude this survey by briefly describing three ongoing lines of NMT research. At first glance they may seem unrelated to the domain adaptation scenarios described above. However, all have seen successful application of domain adaptation techniques in recent years. 
In each case we briefly describe  the problem, and assess whether it meets the criteria for a single-domain or multi-domain adaptation problem using the points summarized in Section 2. We then summarize recent approaches to each problem, focusing on those that use a domain adaptation framing. Some of these approaches have been mentioned elsewhere in this survey: here we put them in the context of their broader line of research. We do not intend this to be a comprehensive survey of work on each problem, and refer to prior surveys  where available in each case. Instead we include this section  as  a demonstration of the  power and wide applicability of domain adaptation techniques, as well as guidance for applying the (multi-)domain adaptation framing to new problems. 

In this section we demonstrate the general utility of techniques in this survey for notions other than changing text topic or genre. The `domains' referred to in this section are not necessarily implicit given the source sentence in the same way that topic or genre might be.  However, we note that even for traditional domain adaptation we often handle sentences where domain is not implicit. Indeed, identical sentences may be  translated in multiple different ways depending on target domain, whether due to lexical ambiguity, customer-specific terminology or genre-appropriate formality. Any of these could be resolved by controlling the target domain. Likewise, we may wish to control an output language, entity gender or document `domain' despite not being able to infer it from the source. 

\subsection{Low Resource Language Translation as a Domain Adaptation Problem}
Neural machine translation systems have achieved impressive performance on language pairs with millions of well-aligned bilingual training sentence pairs available. For many language pairs, such resources are not available. A survey on  natural language processing in such low resource cases is given in \shortciteA{hedderich-etal-2020-lowresource}. 

Faced with machine translation between a low resource language pair, we may wish to improve the available resources, for example by generating synthetic bilingual data. Alternatively we may decide to minimize the need for additional resources, for example by adapting an existing NMT model trained to translate from the same source language. These approaches are closely related to those described for domain adaptation in this article: in effect, they treat the low resource language pair as its own domain.

A related approach to low resource language translation is multilingual translation. In multilingual NMT, the goal is to translate to and/or from more than one language with a single model \shortcite{dong-etal-2015-multi,aharoni-etal-2019-massively}. Multilingual NMT is a broad field of research which does not necessarily involve low resource translation. However we are particularly interested in the application of multilingual NMT to low resource languages  \shortcite{mueller-etal-2020-analysis}. This can be seen as analogous to a multi-domain adaptation scenario: we wish to achieve good performance on multiple distinct sets of text without compromising translation on any, and ideally with the distinct `domains' benefiting each other.

\subsubsection{Can Low Resource Language Translation Be Treated as a Domain Adaptation Problem?}
Low resource language translation can be seen as a single-domain adaptation problem in most cases, although for multilingual NMT systems it may become a multi-domain adaptation problem:

\begin{itemize}
\item We wish to improve translation performance on sentences from a low-resource language pair. These should be very identifiable.
\item We wish to avoid retraining, since we know that training from scratch on the low-resource language pair alone is unlikely to be effective.
\item Training a system to translate between only a single pair of languages will not be a multi-`domain' problem. However, if we are training a one- or many-to-many multilingual NMT model, we may want to adapt so as to translate into several languages simultaneously. This may mean avoiding forgetting previous language pairs, learning multiple languages simultaneously, or learning new languages in a few-shot manner.
\end{itemize}

\subsubsection{Improving Low Resource Language Translation by Adaptation} 
Early and straightforward data-centric approaches to low resource language translation, like domain adaptation, center around tuning an existing model on data for the new language pair, usually sharing either the target language \shortcite{zoph2016transfer} or source language 
\shortcite{kocmi-bojar-2018-trivial}. A distinction is that in this case catastrophic forgetting of the original model's abilities is less of a concern, since it is likely that a user can pre-determine which languages will be translated with which model. However, many techniques are applicable to both single-language-pair domain adaptation and cross-lingual transfer learning.

If the architecture is to be kept the same, lack of vocabulary overlap can be a difficulty, although less so for related languages or those that can leverage similar BPE vocabularies \shortcite{nguyen-chiang-2017-transfer}. 
 Alternatively, both the data and architecture can change: for example, 
\shortciteA{kim-etal-2019-pivot} explore use of adding or sharing network components for transfer learning new languages.  


Monolingual data is often more freely available than bilingual data for low resource translation, as for small-domain translation. Data-centric approaches to low-resource language translation may therefore adapt to semi-synthetic datasets created by forward- or back-translation \shortcite{rubino-etal-2021-extremely,karakanta-etal-2018-without}. Even if this pseudo-parallel data is generated using a relatively weak low resource translation system, it still may be beneficial for further tuning that system \shortcite{currey-heafield-2019-zero}. An alternative option is to generate purely synthetic data: for example
\shortciteA{fadaee-etal-2017-data} target rare words for low resource NMT by situating them in new, synthetic contexts.

If tuning a previously-trained model on a new language pair, it is not usually necessary to maintain performance on the original language pair. However, if e.g. the original source language is shared with the new language pair, it may be better to avoid overfitting the original encoder on the new dataset. To this end, previous work has experimented with freezing some parameters \shortcite{ji2019cross}, or jointly training on the low and high resource language for `similar language regularization' \shortcite{neubig-hu-2018-rapid}. It may also be possible to leverage monolingual data from the low resource language during tuning, for example to regularize adaptation \shortcite{baziotis-etal-2020-language}. Such approaches are clearly reminiscent of domain adaptive tuning techniques described in section \ref{litreview-domainforgetting}.  

A separate approach develops multilingual NMT models, which may translate between very many language pairs. As for single-low-resource-language adaptation, most `domain'-adjacent techniques used here are naturally data-centric. For example, if the language set includes both high and low resource language pairs, these may act analogously to high and low resource domains, with lower resource language pairs benefiting from the higher resource lexical representations \shortcite{gu-etal-2018-universal}.  Multilingual models may also benefit from pseudo-parallel datasets,  even for language pairs with no parallel data \shortcite{firat-etal-2016-zero}. New language pairs may be introduced at different points during training, with language or language pair tags, or subsets of the model parameters reserved for individual language pairs \shortcite{blackwood-etal-2018-multilingual,zhang-etal-2020-improving}. In terms of architecture-centric approaches, the adapter scheme described in section \ref{sec:adapter} was proposed simultaneously for multi-domain adaptation and multilingual NMT \shortcite{bapna-firat-2019-simple,philip-etal-2020-monolingual}, highlighting the conceptual connection between these lines of research.

\subsection{Gender Bias in Machine Translation as a Domain Adaptation Problem}
\label{sec:gender}
Translation into languages with grammatical gender involves correctly inferring the grammatical gender of all entities in a sentence. In some languages this grammatical gender is dependent on the social gender of human referents. For example, in German, translation of the entity `the doctor' would be feminine for a female doctor -- \emph{Die \"Arztin} -- or masculine for a male doctor -- \emph{Der Arzt}. 

\begin{table}[ht]
    \centering
\begin{tabular}{|p{3cm}|p{11.7cm}|}
\hline
English source & The \textbf{doctor} helps the patient, although  \textbf{she} is busy\\
\multirow{2}{*}{German reference} & \textbf{Die \"Arztin} hilft dem Patienten, obwohl \textbf{sie} beschäftigt ist \\
& \textit{The [female] doctor helps the [male] patient, although she is busy} \\
\multirow{2}{*}{\parbox[t]{3cm}{MT with a bias-related mistake}}  & Der Arzt  hilft \textbf{der Patientin}, obwohl \textbf{sie} beschäftigt ist\\
& \textit{The [male] doctor helps the [female] patient, although she is busy} \\
\hline
English source & The \textbf{nurse} helps the patient, although  \textbf{he} is busy\\
\multirow{2}{*}{German reference} & \textbf{Der Krankenpfleger} hilft dem Patienten, obwohl \textbf{er} beschäftigt ist \\
& \textit{The [male] nurse helps the [male] patient, although he is busy} \\
\multirow{2}{*}{\parbox[t]{3cm}{MT with a bias-related mistake}}  & Die Krankenschwester  hilft \textbf{dem Patienten}, obwohl \textbf{er} beschäftigt ist\\
& \textit{The [female] nurse helps the [male] patient, although he is busy} \\
\hline
\end{tabular}
\caption{Examples of  mistranslation relating to gender bias effects. Bolded words are entities inflected to correspond to the pronoun. Machine translations from Google Translate 03/21.}
\label{tab:genderbias-ex}
\end{table}

In practice, however, many NMT models struggle to generate such inflections correctly \shortcite{prates2019assessing}. Gender-based errors are particularly common when translating sentences involving coreference resolution. Table \ref{tab:genderbias-ex} gives two typical examples. \shortciteA{stanovsky-etal-2019-evaluating} explore these mistakes and demonstrate that they tend to reflect social gender bias: machine translation tends to translate based on profession-based gender stereotypes instead of correctly performing coreference resolution and translating using this meaningful context. This may be because the systems are influenced by the higher frequency of masculine-inflected doctors and feminine-inflected nurses in training data, resulting from historical or cultural imbalances in the society that produces this data. 

Such effects are commonly referred to as  systems exhibiting gender bias. Interestingly, the effects can be interpreted in much the same way as a traditional text domain: certain gendered terms have become associated with masculine or feminine grammar in the target language. The model effectively derives a `gender' domain implicitly from even gender-ambiguous source sentences (e.g. `the doctor helps the patient'), even though this is not desirable. Mitigation techniques typically revolve around controlling this `domain' more sensibly. \shortciteA{savoldi-bias-review}  review these effects and related work in NMT.


\subsubsection{Can Gender Bias be Treated as a Domain Adaptation Problem?}
The problem of mitigating the effects of gender bias in NMT can be cast as a multi-domain adaptation problem:

\begin{itemize}
\item We wish to improve translation for sentences with a distinct  vocabulary distribution, which can be interpreted as a domain: sentences containing gendered terms which do not match existing social biases, such as female doctors and male nurses.
\item We wish to avoid retraining: biases may be reinforced by the overall training set.
\item We want to translate at least two domains. One consists of sentences affected by bias, the other of sentences that are unaffected -- we do not wish to compromise translation of sentences without gendered terms. We may also want to continually adapt to  new `domains' to  mitigate newly identified biases, since we are unlikely to successfully pre-determine all relevant biases and their sources.
\end{itemize}

\subsubsection{Mitigating the Effects of Gender Bias By Adaptation}

While there has been interest in mitigating undesirable data bias prior to training, this may be not be practical in terms of computation or in terms of ability to adjust the data \shortcite{tomalin-etal-2020-rethinking}. Such approaches may also not be conceptually sensible, as there are countless  ways in which biases could conceivably manifest in generated natural language,  relating to gender or otherwise \shortcite{hovy-etal-2020-sound,shah-etal-2020-predictive}, so speaking in terms of simple biases or imbalances that can be addressed is not clearly meaningful. Instead, a range of approaches benefit from treating gender handling as a domain.

Improving gender translation by adapting to new data has been demonstrated in several ways, although as with domain adaptation these approaches generally involve some mitigation of forgetting and/or overfitting during training or inference. \shortciteA{jwalapuram-etal-2020-pronoun} adapt, using a discriminative loss function to mitigate overfitting, to sentences containing pronouns that were previously mistranslated. \shortciteA{saunders2020genderbias} adapt to semi-synthetic forward-translated gendered sentences and fully synthetic template-generated sentences, exploring adaptation with EWC and constrained retranslation to avoid forgetting.  \shortciteA{costa-jussa-de-jorge-2020-fine} fine-tune on natural gender-balanced data, and mitigate forgetting by mixing in a proportion of general-domain data.

The idea of controlling machine translation gender inflections with some form of tag, reminiscent of domain tagging, has been proposed in several forms. \shortciteA{vanmassenhove-etal-2018-getting} incorporate a `speaker gender' tag into all training data, allowing gender to be conveyed at the sentence level.  More relevant to domain adaptation, explicit gender tags have been introduced during model adaptation \shortcite{saunders2020neural}. Interestingly, implicit gender tags in the form of gendered `prompts' such as prepending the phrase `he said' or `she said' have been demonstrated by \shortciteA{moryossef-etal-2019-filling} to  control translation gender at inference time, reminiscent of `priming' techniques described in section \ref{sec:prepostprocess}. This approach effectively provides gender domain tags without requiring any changes to the model.

\subsection{Document-Level Translation as a Domain Adaptation Problem}
\label{sec:docnmt}
Document-level machine translation,  summarized in section \ref{sec:doc}, has come to refer to two connected ideas. The first is translating one specific document, accounting for its terminology or writing style. The second is translation of documents in general, accounting for context beyond the individual sentence - for example to resolve pronoun coreference when handling gender, as in section \ref{sec:gender}. Document-specific terminology can  benefit from extra-sentential context, as shown in Table \ref{tab:document-ex}, so the ideas can be considered linked, and are often explored together in the literature -- a thorough survey of which can be found in \shortciteA{maruf2021survey}.  Translating a specific document can clearly be cast as an adaptation problem. We  find that incorporating document context into an existing model often also involves adaptation-adjacent approaches, since document-aligned bilingual datasets are usually much smaller than sentence-aligned bilingual datasets \shortcite{voita-etal-2019-good}.

\begin{table}[t]
    \centering
\begin{tabular}{|p{3.8cm}|p{10cm}|}
\hline
English & \emph{[X]} is a portrait photographer$_{12}$. She$_1$ is known for shooting$_2$ in the woods.\\
German translations of individual sentences & \emph{[X]} ist \textbf{Porträtfotograf}. \text{Sie} ist bekannt dafür, im Wald zu \textbf{schießen}. \\
& \emph{X is a [male] portrait photographer. She is known for shooting [a weapon] in the woods.}\\
German translations with context & \emph{[X]}  ist \textbf{Porträtfotografin}. Sie ist bekannt dafür, im Wald zu \textbf{fotografieren}.\\
& \emph{X is a [female] portrait photographer. She is known for taking photographs in the woods.}\\

\hline
\end{tabular}
\caption{Example of English-to-German translation improved by document context. Two terms with their cross-sentence resolution are numbered in the English sentence: 1 involves an anaphoric pronoun and 2 is a case of lexical ambiguity. Ambiguous German translations (bolded) are resolvable given document context.}
\label{tab:document-ex}
\end{table}

\subsubsection{Can Document NMT Be Treated as a Domain Adaptation Problem?}
Improving document-level translation can be seen as a domain adaptation problem, and may be seen as a multi-domain adaptation problem depending on efficiency requirements:

\begin{itemize}
\item We wish to improve translation performance on (at least) one document, which may use specific terminology  or a distinct writing style.
\item We wish to avoid retraining from scratch: an individual document or  set of document-aligned data is very small compared to typical sentence-aligned generic training sets, so retraining would be especially inefficient.
\item  If we are not willing to tune one model per document, we may wish to use the same model to translate well across multiple documents. This would be a multi-domain adaptation problem for the sake of efficiency. However, it is perhaps more likely that only one document is of interest at a time.
\end{itemize}

\subsubsection{Improving Document-Level Translation By Adaptation}

Many of the data-centric adaptation approaches described in section \ref{litreview-dataselection} use some `seed' text to extract related bilingual or monolingual data. These can be used for document-level translation with the document to be translated as a seed. An example of this approach is \shortciteA{kothur-etal-2018-document}, who adapt to a lexicon containing novel words in a test document.

Attempts to use extra-sentential context in document translation have primarily incorporated additional sentences into the model input as described in section \ref{sec:doc}. However, an alternative approach reminiscent of domain adaptation is to condense document information into tag or label form (\ref{sec:doctag}). \shortciteA{jehl-riezler-2018-document} use inline document-content tags for patent translation, integrating information at the word and sentence-level. \shortciteA{DBLP:journals/corr/abs-2004-14927} show that incorporating document context can improve domain-specific translation, suggesting that domain tags and document tags may function similarly. 
\shortciteA{kim-etal-2019-document} demonstrate that contextual and lexical information can be incorporated into a very minimal form, reminiscent of tagging approaches, rather than necessitating encoding entire additional sentences or documents. 

The neural network architecture may need to be adjusted in order to incorporate context beyond the individual sentence. A typical change is simply adding additional context source sentences alongside additional encoder parameters. \shortciteA{stojanovski-fraser-2019-combining} take this approach, initialising the new context parameters randomly in a pre-trained model before tuning on smaller document-level datasets. \shortciteA{voita-etal-2019-good}  initialise and tune a context-aware decoder on a relatively small amount of document-sensitive data. 
\shortciteA{ul-haq-etal-2020-improving} first train a generic NMT model, then add and adapt context-sensitive hierarchical attention networks on document-specific data. 
A final set of domain-adaptation-adjacent approaches to document-level translation incorporates document information at inference time (section \ref{sec:review-inference}):  \shortciteA{voita-etal-2019-context} carry out monolingual document repair with a specialised system trained on `in-domain' -- context-aware -- data. \shortciteA{stahlberg-etal-2019-cued} similarly rerank translations with a language model that incorporates document context.

\section{Conclusions and Future Directions}
\label{sec:conclusions}

Domain adaptation lets NMT models achieve good performance on language of interest with limited training data, and without the cost of retraining the model from scratch. Adaptation may even allow better performance than from-scratch training on a given domain. Ongoing challenges for machine translation can be framed as domain adaptation problems, inviting the application of the well-tested techniques reviewed in the earlier sections of this article.  As our case studies in section \ref{litreview-casestudy} have shown, there is already a trend towards using adaptation for pseudo-domains other than straightforward provenance, topic or genre. It remains to be seen which other lines of  NMT research might benefit from domain adaptation techniques. 

We conclude this survey with a view to the future, highlighting four areas that are of particular interest for future work on NMT domain adaptation: extremely fine-grained adaptation,  unsupervised adaptation,  efficiency,  and intentional forgetting. We have touched on these at various points in this survey; here we emphasise their relevance to future  research.

As observed in section \ref{litreview-domaindescription}, text domains can be difficult to interpret, with topic and style not necessarily consistent even within one document. Fine-grained adaptation has grown in popularity as researchers seek to compensate by reducing the granularity of data selection and tuning. Models may be tuned for a particular user
\shortcite{michel-neubig-2018-extreme,buj-etal-2020-nice}, or for every sentence \shortcite{farajian-etal-2017-multi,li-etal-2018-one,mueller-lal-2019-sentence}. While fine-grained adaptation may improve lexical choice, it is often slow, and risks overfitting to irrelevant data \shortcite{li-etal-2018-one}. Future work in fine-grained or `extreme' adaptation may therefore focus on extraction of relevant data (section \ref{litreview-dataselection}) or ways to adapt at a per-sentence level without changing the model at all (section \ref{sec:prepostprocess}).

Unsupervised domain adaptation, which tunes a system without access to parallel data, is also growing in popularity. A low resource domain -- especially a very fine-grained domain --  may not have parallel corpora for a particular language pair. However, in-domain \emph{monolingual} data is more widely available, and can let us produce pseudo-parallel data by forward or back translation (section \ref{sec:syntheticdata}). Open questions involve the best ways of extracting large amounts of in-domain data from multiple corpora given small amounts of in-domain data, or  of producing pseudo-parallel in-domain data from monolingual in-domain data. With interest in adapting to ever-smaller domains, sufficient in-domain parallel data is increasingly scarce: we predict interest in unsupervised adaptation will correspondingly grow.


Efficiency concerns stem from the tension between high-quality, fine-grained adaptation and practical use of limited resources.  As  typical NMT models grow larger \shortcite{wei-etal-2020-multiscale} and more work emphasises extreme adaptation at the user or sentence level, there is growing awareness that training and storing large numbers of  deep neural models is unsustainable from both a financial and an environmental perspective \shortcite{strubell-etal-2019-energy,bender2021dangers}. Energy use and carbon footprint are certainly greater when training from scratch on hundreds of millions of sentences as compared to fine-tuning models only on relevant data. However, it may be that the aggregate effect of `personalizing' large NMT systems for many users   outweighs the reduced per-model training time. In general, we predict that efficiency in terms of tuning and storing models will continue to grow in importance. This convergence of requirements may draw attention to approaches which adapt using fewer tuning steps, or do not adapt the entire model. These  may be variations on adapter layers (section \ref{sec:adapter}), ways to find sparse or underused subsets of parameters for minimal model adaptation (section \ref{sec:freeze}), or means of adapting the output that avoid  parameter tuning completely (section \ref{sec:prepostprocess}).


Finally, we note an interesting trend of machine learning research into neural \emph{unlearning} \shortcite{DBLP:conf/amcis/KwakLPL17,DBLP:conf/ccs/CaoYASMY18,DBLP:conf/alt/Neel0S21}. We have previously described catastrophic forgetting as a pitfall for adaptation, but forgetting can be viewed as a feature, not a bug. As we saw in case studies on low resource and gender translation, domain adaptation techniques can find application in intentionally forgetting or abandoning certain behaviour which is no longer desirable, rather than specifically learning new behaviour.  We might want to unlearn domains we know we will not translate in the future, or abandon a tendency to default to outdated translations, or mitigate more subtle effects resulting from word associations relating to demographic biases.  Adaptation for the purposes of intentional forgetting may become more relevant with new risks connected to NLP systems. For example, recent research has suggested that neural language models can be `attacked' to retrieve their training data \shortcite{wallace-etal-2020-imitation,DBLP:journals/nn/GongPXQT20}. Given this risk the ability to redact specific translation examples via unlearning might be desirable, especially in privacy-sensitive cases such as biomedical translation. As a relatively new area of research, whether unlearning  can improve NMT remains an open question.


Natural language is both complex and evolving, as are the AI systems that interact with natural language. With this survey we hope to draw attention to the possible benefits and drawbacks of different approaches to domain adaptive translation, as well as their possible applications. We hope that future work on adapting neural machine translation will focus not only on individual domains of immediate interest, but on the range of machine translation abilities that we wish to maintain or abandon.


\bibliography{refs}
\bibliographystyle{theapa} 
\end{document}